\documentclass[11pt]{article}

\usepackage[final]{acl}

\usepackage{times}
\usepackage{latexsym}

\usepackage[T1]{fontenc}
\usepackage[utf8]{inputenc}

\usepackage{microtype}

\usepackage{inconsolata}

\usepackage{graphicx}
\usepackage{amsmath}
\usepackage{booktabs, xcolor, array, multirow, listings}  
\usepackage{adjustbox}
\usepackage{tcolorbox, amssymb}
\usepackage{pifont}  
\newcommand{\cmark}{\ding{51}} \newcommand{\xmark}{\ding{55}}
\usepackage[T1]{fontenc}
  \usepackage[utf8]{inputenc}
\title{When Users Don't Ask: \\Benchmarking Context-Driven Memory Retrieval in Conversational Agents}

\author{Wen-Yu Chang \quad Yun-Nung Chen\\
         National Taiwan University, Taipei, Taiwan\\
        \texttt{f10946031@csie.ntu.edu.tw}\quad \texttt{y.v.chen@ieee.org}
         }

\begin{document}
\maketitle
\begin{abstract}
Large language models (LLMs) are increasingly deployed as long-horizon
conversational agents, motivating growing interest in memory systems. However,
existing benchmarks primarily evaluate memory through QA-style probing rather
than in-situ conversational usage. We introduce \textbf{\textsc{LoCoMo-Conv}}, a
conversational memory benchmark derived from LoCoMo with four query styles:
\emph{dialog}, \emph{implicit}, \emph{counterfactual}, and \emph{composed}.
Across five representative memory systems, we evaluate both retrieval recall and
end-to-end response quality. Our experiments show that conversational framing
exposes substantial retrieval gaps overlooked by QA benchmarks, especially on
implicit and composed queries, which multi-facet query rewriting narrows for raw-turn memory but not abstractive memory. We further find that strong retrieval does
not fully translate into response quality, and that implicit queries exhibit
\emph{silent grounding}, where memory improves contextual grounding without
explicitly surfacing the gold fact. These results point to reasoning-based
memory elaboration as a promising direction, and we release auxiliary
\texttt{supportive\_memory} annotations capturing conversationally useful
context beyond the original gold evidence.\footnote{\url{https://github.com/MiuLab/LoCoMo-Conv/}}
\end{abstract}
\begin{table*}[t]
\centering
\small
\setlength{\tabcolsep}{8pt}
\renewcommand{\arraystretch}{1.1}
\begin{tabular}{@{}l l l c@{}}
\toprule
\textbf{Benchmark} & \textbf{Query form} & \textbf{Evaluation} & \textbf{Mem.\ sys.} \\
\midrule
LoCoMo~\shortcite{maharanaEvaluatingVeryLongTerm2024} & 3rd-person QA & QA match & \cmark \\
LongMemEval~\shortcite{wu2025longmemeval} & 3rd-person QA & QA match & \cmark \\
PersonaMem (v1/v2)~\shortcite{jiangPersonaMemv2PersonalizedIntelligence2025} & 1st-person, in-situ & multiple choice & \xmark \\
MADial-Bench~\shortcite{he-etal-2025-madial} & emotion-support dialogue & generation & partial \\
LoCoMo-Plus~\shortcite{li-etal-2026-locomo} & conversation continuation & constraint consistency & \cmark \\
AMemGym~\shortcite{jiayang2026amemgym} & simulated interaction & structured metrics & \cmark \\
MemoryAgentBench~\shortcite{hu2026evaluating} & 3rd-person QA & QA match & \cmark \\
\midrule
\textbf{LoCoMo-Conv (ours)} & \textbf{4 1st-person conv.\ styles} & \textbf{retrieval + free-form response} & \cmark \\
\bottomrule
\end{tabular}
\caption{Comparison with existing memory benchmarks. LoCoMo-Conv varies only the query framing over fixed LoCoMo memories and scores retrieval and response separately.}
\label{tab:benchmark_positioning}
\end{table*}

\begin{table*}[t] 
\centering
\small
\renewcommand{\arraystretch}{1.3}
% 這裡加入了 >{\raggedright\arraybackslash}，可以徹底修復 \raggedright 導致 \\ 失靈的問題
\begin{tabular}{@{} >{\raggedright\arraybackslash}p{0.48\textwidth} | >{\raggedright\arraybackslash}p{0.48\textwidth} @{}} 
\toprule
\multicolumn{1}{c}{\textbf{Knowledge Context (Original \& Memory)}} & \multicolumn{1}{|c}{\textbf{Conversational Query Styles}} \\
\midrule 

% === 左半邊 ===
\textbf{Original Question:} \newline \emph{What were Deborah's mother's hobbies?} \vspace{0.5em} \newline
\textbf{Gold Answer:} \newline reading, traveling, art, cooking \vspace{0.5em} \newline
\textbf{Memory:} \newline
$\bullet$ \texttt{[D2:17]} Deborah: ``she'd sit there every night with a book and a smile...'' \newline
$\bullet$ \texttt{[D2:19]} Deborah: ``Travel was also her great passion!'' \newline
$\bullet$ \texttt{[D12:3]} Deborah: ``My mom was interested in art\ldots'' \newline
$\bullet$ \texttt{[D29:7]} Deborah: ``My mom had a big passion for cooking\ldots'' 
& % 換到右欄
% === 右半邊 ===
\textbf{Dialogue Question:} \newline
\emph{Do you remember what I told you about my mom's hobbies?} \vspace{0.8em} \newline
\textbf{Implicit User Turn:} \newline
\emph{I'm trying to think of a meaningful birthday gift for my mom, but I'm totally drawing a blank on what she'd actually enjoy.} \vspace{0.8em} \newline
\textbf{Counterfactual User Turn:} \newline
\emph{I was chatting with a coworker today about my mom, and I mentioned how she spent all her time \textcolor{red}{gardening and knitting}---I think that's what I told her, right?} \\ % 這裡的 \\ 現在能正確結束整列了

\bottomrule 
\end{tabular}
\caption{One LoCoMo QA rewritten into our 3 conversational query styles. The gold answer and gold evidence turns (left) are shared across all rewrites; only the user-side phrasing (right) changes. The counterfactual rewrite injects an incorrect premise (\textcolor{red}{red}) that the assistant must correct.}
\label{tab:query_style_examples}
\end{table*}

% ============================================================
% Table 3: Composed example (multi-memory)
% ============================================================
\begin{table*}[t]
\centering
\small
\renewcommand{\arraystretch}{1.2}
\setlength{\tabcolsep}{6pt}
\begin{tabular}{@{}p{0.47\textwidth}|p{0.47\textwidth}@{}}
\toprule
\multicolumn{2}{c}{\textbf{Example of Composed Query}} \\
\midrule
\multicolumn{1}{c|}{\textbf{Source QA\textsubscript{1}}} & \multicolumn{1}{c}{\textbf{Source QA\textsubscript{2}}} \\
\midrule
\textbf{Q:} \emph{What helped Deborah find peace when grieving deaths of her loved ones?}
& \textbf{Q:} \emph{Why did Deborah spend time in the garden?} \\
\textbf{Gold Answer:} yoga, old photos, the roses and dahlias in a flower garden, nature
& \textbf{Gold Answer:} to find comfort after losing a friend \\
%\midrule
\textbf{Memory:} & {\textbf{Memory:}} \\
%\midrule
\texttt{[D1:15]} Deborah: ``Yoga helped me find peace during a rough time, and now I'm passionate about sharing that with others.''
\newline\texttt{[D2:3]} Deborah: ``\ldots it's comforting to look back on the great memories. We looked at the family album. Photos give me peace during difficult times.''
\newline\texttt{[D6:4]}$^{\ast}$ Deborah: ``The roses and dahlias bring me peace. I lost a friend last week, so I've been spending time in the garden to find some comfort.''
\newline\texttt{[D15:29]} Deborah: ``Nature helps me find peace every day---it's so refreshing!''
& \texttt{[D6:4]}$^{\ast}$ Deborah: ``The roses and dahlias bring me peace. I lost a friend last week, so I've been spending time in the garden to find some comfort.''
\newline\footnotesize\textcolor{gray}{($\ast$ \texttt{[D6:4]} is shared by both QAs --- the overlap that motivates composing them into one cluster.)} \\
\midrule
\multicolumn{2}{@{}p{0.95\textwidth}@{}}{\textbf{Composed User Turn:} \hfill \footnotesize\textcolor{gray}{(LLM-generated from QA\textsubscript{1} + QA\textsubscript{2})}} \\
\multicolumn{2}{@{}p{0.95\textwidth}@{}}{\emph{``I've been feeling that familiar heaviness again lately, and I'm trying to remember exactly what worked for me last time I was struggling to cope with a loss.''}} \\
\midrule
\multicolumn{2}{@{}p{0.95\textwidth}@{}}{\textbf{Expected synthesis} \hfill \footnotesize\textcolor{gray}{(LLM-generated rubric)}} \\
\multicolumn{2}{@{}p{0.95\textwidth}@{}}{The assistant should surface \emph{multiple} past coping strategies from QA\textsubscript{1} (yoga, old photos, garden, nature) and ideally connect them to the loss context from QA\textsubscript{2}. A response listing only one item, or only acknowledging the user's situation without surfacing past strategies, is judged as partial coverage.} \\
\bottomrule
\end{tabular}
\caption{Unlike dialog/implicit/counterfactual, which rewrite a single LoCoMo QA, \textbf{composed queries combine two source QAs} into one conversational request requiring multi-fact synthesis. The top half shows the source QAs and their evidence turns, with shared turns ($\ast$) marking the cluster overlap. The bottom row shows the LLM-generated composed query and synthesis rubric.}
\label{tab:composed_example}
\end{table*}

\section{Introduction}
% Large Language Models (LLMs) have rapidly evolved from single-turn conversational assistants into reasoning-capable agents capable of executing complex multi-step tasks. The emergence of agentic platforms and harnesses, such as  OpenClaw, Hermes Agent and Anthropic Claude Cowork, has further positioned LLMs as persistent everyday assistants with which users interact continuously over extended periods, rather than through isolated and independent sessions. Consequently, users increasingly expect these systems to behave as persistent entities that retain and utilize information from prior interactions, instead of functioning as stateless chatbots that reset at the beginning of each session. This transition from session-isolated interactions to cross-session shared context introduces fundamentally new challenges and requirements for the underlying memory system.~\citep{caoHiGMemHierarchicalLLMGuided2026,chhikaraMem0BuildingProductionReady2025,fangLightMemLightweightEfficient2026,packerMemGPTLLMsOperating2024,kangMemoryOSAI2025,huMemoryAgeAI2026,shenAnchorMemAnchoredFacts2026,xuAMEMAgenticMemory2025,zhongMemoryBankEnhancingLarge2023}
\begin{figure}[!t]                                                                                                       
  \centering                                                                                                               
  \includegraphics[width=\columnwidth]{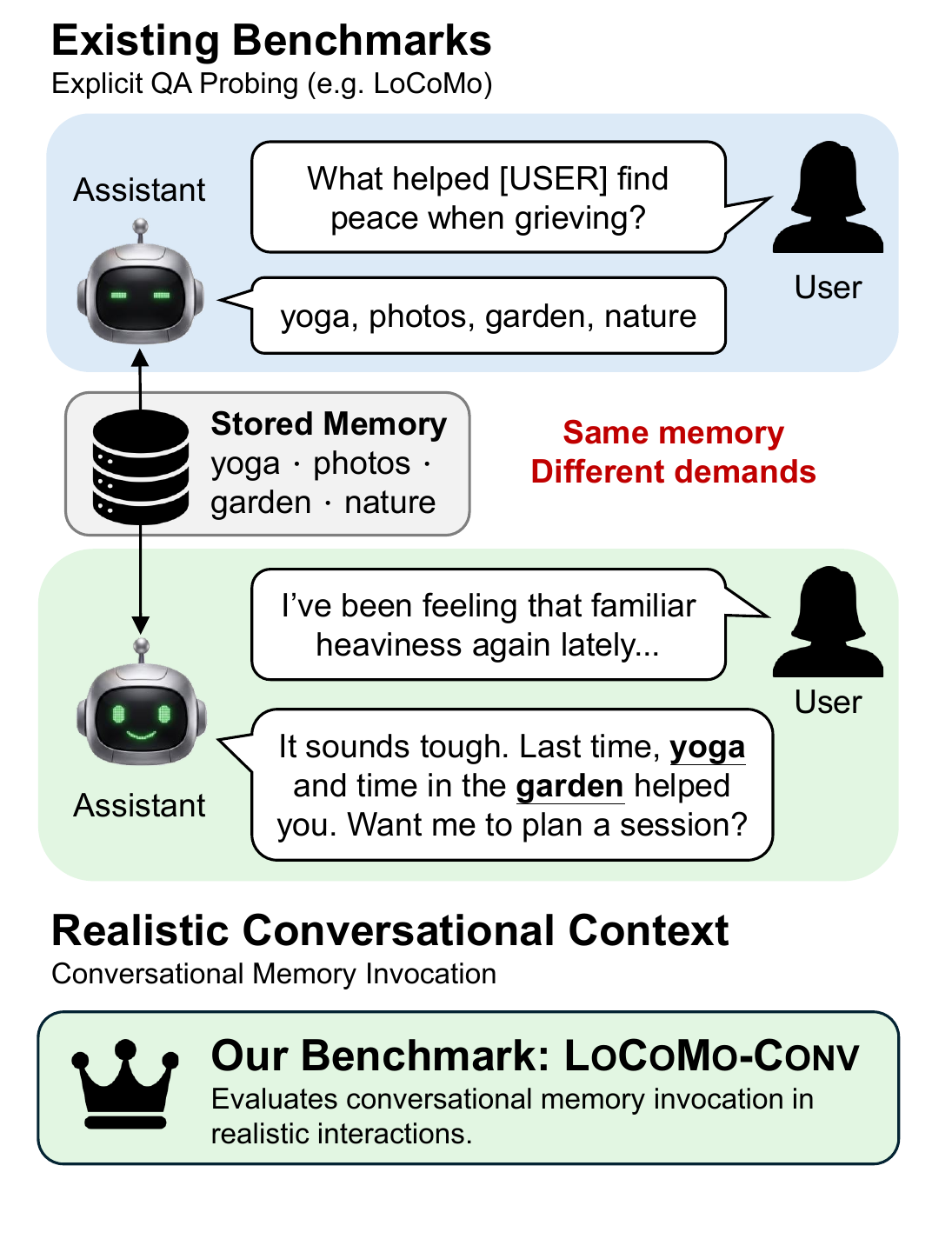}
  \caption{Illustration of the gap between \textsc{LoCoMo-Conv} and the existing benchmarks}                                                                                                            
  \label{fig:benchmark_gap}                                                                                              
  \end{figure} 
  Large language models (LLMs) are increasingly deployed as long-horizon conversational agents that users interact with across sessions. This shift makes memory a core requirement: assistants must not only store prior interactions, but also retrieve and use them appropriately when a future conversation calls for them. Recent work has proposed diverse memory architectures and benchmarks for evaluating long-term conversational memory. However, most existing evaluations still probe memory through explicit QA-style queries (as shown in Table~\ref{tab:benchmark_positioning}), leaving open whether these systems can support natural conversational interaction.

To address this gap, we introduce \textbf{LoCoMo-Conv}, a conversational memory
benchmark that recasts LoCoMo's QA pool into four query styles reflecting how
users naturally invoke memory---\textit{dialog}, \textit{implicit},
\textit{counterfactual}, and \textit{composed}---and evaluates how
memory-augmented agents use retrieved memory within natural dialogue rather than
under explicit probing.

Across five representative memory systems, we report three findings. First,
conversational framing uncovers substantial retrieval failures masked by
QA-style evaluation, especially on implicit and composed queries; multi-facet
query rewriting narrows the gap, but only for raw-turn memory, not abstractive
memory. Second, strong retrieval does not guarantee grounded responses---
abstractive compression in particular aids matching while discarding the detail
needed to ground a response---pointing to reasoning-based memory elaboration
over lossy compression. Third, we identify \emph{silent grounding}, where memory
improves implicit-query responses even without surfacing the gold fact, exposing
a limitation of strict fact-recall metrics.
\noindent Our main contributions are:
\begin{enumerate}
\item We introduce \textsc{LoCoMo-Conv}, a conversational benchmark with four
first-person query styles that evaluates memory systems beyond QA-style probing.

\item We provide a unified evaluation framework spanning both retrieval and
response generation, enabling systematic comparison of extractive and
abstractive memory systems under conversational settings.

\item We release \texttt{supportive\_memory}, an auxiliary annotation layer
capturing conversationally supportive context beyond the original gold
evidence, facilitating the analysis of memory grounding beyond explicit fact
recall.
\end{enumerate}
% \noindent Our main contributions are:
% \begin{enumerate}
% \item We introduce \textsc{LoCoMo-Conv}, extending \textsc{LoCoMo} with four
% conversational query styles to expose memory gaps hidden by QA-style evaluation.
% \item We jointly evaluate five systems across retrieval and response quality,
% showing that strong retrieval does not yield grounded responses, and that the
% effectiveness of query rewriting and the cost of abstractive compression both
% depend on the memory representation.
% \item We identify \emph{silent grounding} and release
% \texttt{supportive\_memory}, an auxiliary annotation layer capturing
% conversationally supportive context beyond the original gold evidence.
% \end{enumerate}
\section{Related Work}
\subsection{Benchmarks for Long-Term Conversational Memory}
\label{sec:related_benchmarks}

Existing long-term memory benchmarks differ mainly in how memory is probed and what is scored. \textbf{LoCoMo}~\citep{maharanaEvaluatingVeryLongTerm2024}
pioneers very long multi-session conversations with third-person QA across five categories
(single/multi-hop reasoning, temporal, adversarial, open commonsense).
\textbf{LongMemEval}~\citep{wu2025longmemeval} defines five memory abilities---information
extraction, multi-session reasoning, temporal reasoning, knowledge updates, and abstention---probed
through curated QA over scalable dialogue histories.
\textbf{MemoryAgentBench}~\citep{hu2026evaluating} reformulates long-context datasets into
incremental multi-turn streams to evaluate four memory competencies (accurate retrieval, test-time
learning, long-range understanding, conflict resolution) at scales up to 1.4M tokens. Other
benchmarks target specialized memory failure modes:
\textbf{HaluMem}~\citep{chenHaluMemEvaluatingHallucinations2026} decomposes memory hallucination
into extraction, updating, and QA stages; \textbf{PrefEval}~\citep{zhaoLLMsRecognizeYour2025}
evaluates whether LLMs adhere to user preferences in long-context dialogue; and
\textbf{HorizonBench}~\citep{liHorizonBenchLongHorizonPersonalization2026} probes long-horizon
personalization under evolving preferences with $\sim$163K-token, 6-month conversation histories.

Closer to conversational use, \textbf{PersonaMem}~\citep{jiang2025knowmerespondme} and its
successor \textbf{PersonaMem-v2}~\citep{jiangPersonaMemv2PersonalizedIntelligence2025} target
implicit \emph{user preferences} accumulated across sessions, finding that frontier LLMs achieve
only 37--48\% accuracy on implicit personalization; both, however, evaluate via multiple-choice
selection and probe whether the agent has internalized user traits over time rather than whether
it surfaces relevant memory from a single utterance at query time.
\textbf{MADial-Bench}~\citep{he-etal-2025-madial} evaluates memory-augmented dialogue generation
with proactive recall, but within a single emotion-support domain and with small-scale human
evaluation. \textbf{LoCoMo-Plus}~\citep{li-etal-2026-locomo} tests whether a conversational
continuation stays consistent with latent constraints, where a valid response need not retrieve or
express any specific fact. \textbf{AMemGym}~\citep{jiayang2026amemgym} evaluates memory through
on-policy interaction with simulated users, which is complementary to our fixed-history setting.

Despite this diversity in measurement axes, nearly all existing benchmarks evaluate through
structured probing---third-person QA, multiple-choice selection, constraint checks, or
operation-level traces---rather than free-form conversational response generation, and none
requires an external memory system to retrieve and ground a specific, verifiable fact from a
user utterance that does not ask for it. LoCoMo-Conv departs from this convention by evaluating
\emph{how an assistant integrates memory into a conversational response}, the form in which
memory is actually consumed in deployment.
\subsection{Memory-Augmented Conversational Agents}
\label{sec:related_systems}

Recent work has explored diverse architectural strategies for equipping LLMs with long-term memory in conversational settings. \textbf{mem0}~\citep{chhikaraMem0BuildingProductionReady2025} and \textbf{AnchorMem}~\citep{shenAnchorMemAnchoredFacts2026} extract atomic facts from raw interactions and organize them into structured stores, with AnchorMem further constructing an associative event graph to capture cross-memory dependencies. A second line compresses past interactions into condensed representations: \textbf{COMEDY}~\citep{chenCompressImpressUnleashing2025} eschews retrieval entirely by using a single LLM to generate, compress, and consume compressive memory; \textbf{LightMem}~\citep{fang2026lightmem} adopts a three-stage Atkinson--Shiffrin-inspired pipeline; and \textbf{MemAgent}~\citep{yuMemAgentReshapingLongContext2025} processes long inputs in segments under an overwrite policy optimized via reinforcement learning. A third family draws on operating-system principles---\textbf{MemGPT}~\citep{packerMemGPTLLMsOperating2024} introduces tiered memory hierarchies with explicit paging, \textbf{MemoryOS}~\citep{kangMemoryOSAI2025} extends this with short-, mid-, and long-term storage tiers, and \textbf{HiGMem}~\citep{caoHiGMemHierarchicalLLMGuided2026} adopts a two-level event-turn structure with LLM-guided turn selection. \textbf{A-MEM}~\citep{xuAMEMAgenticMemory2025} treats memory construction itself as agent-driven, building an interconnected knowledge network with dynamic linking and memory evolution following Zettelkasten principles. Finally, \textbf{Memora}~\cite{xia2026memora} proposes a harmonic memory representation that separates abstract memory indices from detailed memory values. Each memory consists of a primary abstraction for indexing, multiple cue anchors for diverse retrieval access, and an uncompressed memory value preserving fine-grained details. For evaluation we select five systems that collectively span the dominant paradigms above (AnchorMem, A-MEM, mem0, Memora and
 a dense-retrieval baseline using \texttt{all-MiniLM-L6-v2}), leaving broader-coverage benchmarking to future work.
\section{LoCoMo-Conv}                     
  \label{sec:benchmark}
We construct \textsc{LoCoMo-Conv} by rewriting each LoCoMo10 question into four
conversational query styles while preserving the original gold answers and evidence
\texttt{dia\_ids}.\footnote{Three annotators validated the rewrites on 40 items per style:
they are natural (4.6--4.8 of 5), preserve the original information need, and invoke the
intended memories; see Appendix~\ref{app:human_annotation}.}
\subsection{Conversational Query Styles}
\label{sec:styles}

\paragraph{Dialog.}
A direct first-person conversational reformulation of the original QA query (e.g., ``Do you remember when I\ldots?''). The underlying factual target remains unchanged, but the phrasing is rewritten to resemble natural dialogue rather than third-person question answering.

\paragraph{Implicit.}
A situational conversational utterance in which the user does not explicitly ask a question. Instead, the assistant must infer that a relevant past memory should be proactively surfaced. For example, a factual QA asking about a parent’s hobbies may be rewritten into a conversational situation involving gift selection or reminiscing.

\paragraph{Counterfactual.}
A first-person conversational query containing an incorrect premise about the gold fact. The assistant is expected to identify and correct the false assumption using the underlying conversational memory.

\paragraph{Composed.}
A multi-memory conversational query requiring synthesis across multiple source QAs and evidence spans. We describe the construction process in \S~\ref{sec:composed}.

For all non-composed styles, we prompt \texttt{GPT-5.4-mini} with the original question, gold answer, and speaker identity, and instruct it to generate conversational rewrites that preserve the original information need while remaining natural and first-person in style. For counterfactual queries, the prompt additionally injects a specific false premise. Example rewrites are shown in Table~\ref{tab:query_style_examples}, and all the prompts for query rewriting are provided in Appendix~\ref{app:rewrite_prompts}.
\subsection{Composed Multi-Memory Clusters}
\label{sec:composed}

Unlike the other conversational styles, composed queries are constructed by combining multiple source QAs into a single conversational request that requires multi-memory synthesis.

For each QA instance $q$, let $E(q) \subseteq \mathcal{D}$ denote its gold evidence dia\_ids. We enumerate all QA pairs $(q_i, q_j)$ within the same sample and retain a pair as a candidate cluster if it satisfies three conditions: (i) overlapping evidence, $E(q_i) \cap E(q_j) \neq \emptyset$; (ii) non-identical evidence, $E(q_i) \neq E(q_j)$; and (iii) non-trivial combined evidence, $|E(q_i) \cup E(q_j)| \geq 2$. These constraints ensure that the two QAs are topically related while still contributing distinct information. We retain all valid clusters under these constraints, yielding 1,069 composed clusters in total (36–198 per conversation).

For each cluster, we prompt \texttt{GPT-5.4-mini} with the source QA pairs and their gold evidence turns, and ask it to generate a natural conversational query that implicitly requires both source answers. The composed gold answer is defined as the set of atomic answers from the constituent QAs, while the gold evidence corresponds to the union of their dia\_ids.

Because composed queries require synthesizing multiple atomic facts, binary correctness is often overly strict. We therefore evaluate composed responses using continuous \emph{atomic-fact coverage} (Section~\ref{sec:eval}), which measures the fraction of atomic gold facts correctly covered by the generated response. An example composed cluster is shown in Table~\ref{tab:composed_example}.

\section{Evaluation Methodology}
\label{sec:eval}

We evaluate memory systems along two complementary dimensions: (1) \textbf{retrieval recall} against the gold dialogue IDs, and (2) \textbf{end-to-end response quality} using style-specific LLM judges.

\subsection{Retrieval Recall}

For each query, we compare the top-$K$ retrieved memories against the gold evidence \texttt{dia\_id}s. A gold turn is considered retrieved if its verbatim text appears within any returned memory (case-insensitive). For abstractive systems that expose source metadata, we additionally match retrieved \texttt{dia\_id}s through metadata fields. Retrieval recall is computed as:
\[
\frac{|\text{retrieved} \cap \text{gold}|}{|\text{gold}|}.
\]

% \subsection{Response Quality Judging}

% Retrieved memories are provided to the answer model to generate responses, which are then evaluated using style-specific binary rubrics with GPT-5.4-mini.

% \begin{itemize}
% \item \textbf{Dialog}: the response correctly conveys the gold fact.
% \item \textbf{Implicit}: the response proactively surfaces the relevant memory while engaging with the user’s situation.
% \item \textbf{Counterfactual}: the response identifies and corrects the user’s incorrect premise.
% \item \textbf{Composed}: the response successfully synthesizes all required atomic facts. We additionally report \emph{atomic-fact coverage}, defined as the fraction of gold atomic facts correctly covered.
% \item \textbf{Adversarial}: we measure hallucination rate, defined as the fraction of responses that incorrectly affirm the adversarial false claim.
% \end{itemize}

% For \textit{implicit} queries, we further decompose evaluation into two independent dimensions: \textbf{fact-grounding} (whether the gold fact is surfaced) and \textbf{situation-tying} (whether the response meaningfully engages with the user’s situation). This decomposition enables finer-grained failure analysis (Section~\ref{sec:analysis}).

% The full judge prompts are provided in Appendix~\ref{app:prompts}.

\subsection{Response Quality Judging}

Retrieved memories are provided to the answer model to generate
responses, which are then evaluated by an LLM judge 
using a style-specific rubric. 

\begin{itemize}
\item \textbf{Dialog} and \textbf{Implicit}: \emph{partial-credit
\texttt{fact\_used}} on a 3-level scale ($1.0$ if the substance of the
gold fact is correctly conveyed---paraphrasing and hedging permitted;
$0.5$ if the response captures the central concept of a multi-item gold
but misses specifics; $0.0$ if the response asserts contradicting
content, gives only vague allusion, or completely omits the fact).

\item \textbf{Counterfactual}: a 3-way \emph{unaware / hedge /
corrected} judge mapped to $0 / 0.5 / 1$. Class A (\emph{unaware})
treats the user's false premise as if it were true; class B
(\emph{hedge}) signals awareness of a mismatch but does not state the
ground-truth fact; class C (\emph{corrected}) states (or clearly
implies) the ground-truth fact regardless of whether it explicitly
points out the user's error.

\item \textbf{Composed}: \emph{atomic-fact coverage}, the fraction of
gold atomic facts the response covers (judged independently per fact).
This avoids the strict all-or-nothing failure of a single
\texttt{fact\_used} call for multi-fact composed clusters.
\end{itemize}

The full judge prompts are provided in Appendix~\ref{app:prompts}.

\section{Experimental Setup}
We evaluate five representative systems: \textbf{AnchorMem}~\citep{shenAnchorMemAnchoredFacts2026}, \textbf{A-MEM}~\citep{xuAMEMAgenticMemory2025}, \textbf{mem0}~\citep{chhikaraMem0BuildingProductionReady2025}, \textbf{Memora}\footnote{We disable Memora's default retrieval threshold (0.4), which otherwise returns nothing for 42--71\% of conversational queries.}~\citep{xia2026memora}  and \textbf{NaiveRAG} built with \texttt{all-MiniLM-L6-v2}. We follow the original implementation settings while unifying the embedding model to \texttt{all-MiniLM-L6-v2} and the backbone LLM to \texttt{gemma-4-31B-it}. All systems retrieve top-$K{=}10$ memories\footnote{AnchorMem retrieves \emph{anchor chunks} rather than individual turns, yielding approximately 12.8 turns on average at top-$K{=}10$.}, which are passed to the same answer-generation model.\footnote{\texttt{temperature}=0, \texttt{max\_tokens}=300.}

For response evaluation, we use \texttt{GPT-5.4-mini} with reasoning enabled as the primary judge, and validate agreement against \texttt{Claude-sonnet-4.5} and \texttt{Qwen3.6-35b-A3B} on a stratified 1,488-item subset (Appendix~\ref{app:judge_validation}).

\newcommand{\up}[1]{\textcolor{green!60!black}{\textbf{$\uparrow$#1}}}
\newcommand{\dn}[1]{\textcolor{red!85!black}{\textbf{$\downarrow$#1}}}

\begin{table*}[t]
\centering
\small
\renewcommand{\arraystretch}{1.05}

\begin{adjustbox}{max width=\textwidth}
\begin{tabular}{lllll|llll}
\toprule
\multirow{2}{*}{\textbf{System / Variant}} &
\multicolumn{4}{c|}{\textbf{Retrieval Recall@10}} &
\multicolumn{4}{c}{\textbf{Response Quality}} \\
\cmidrule(lr){2-5} \cmidrule(lr){6-9}
& Dialog & Implicit & Cf & Comp.
  & Dialog & Implicit & Cf & Comp. \\
\midrule

\multicolumn{9}{l}{\textit{No-memory floor / Oracle ceiling}} \\
No Memory
& --- & --- & --- & ---
& .004 & .067 & .251 & .008 \\

Oracle
& --- & --- & --- & ---
& \it .777 & \it .724 & \it .775 & \it .597 \\

\midrule
\multicolumn{9}{l}{\textit{Baselines (raw turns, no LLM at ingest)}} \\
%BM25 (orig.)
%& .482 & .158 & .492 & .132
%& .462 & .220 & .575 & .147 \\
%~~~~~~~~~~~(+rewrite)
%& .552 \up{.070}
%& .271 \up{.113}
%& .529 \up{.037}
%& .188 \up{.056}
%& .534 \up{.072}
%& .301 \up{.081}
%& .585 \up{.010}
%& .165 \up{.018} \\

%~~~~~~~~~~~(+cot)
%& --- & --- & --- & ---
%& .483 \up{.021}
%& .258 \up{.038}
%& .426 \dn{.149}
%& .183 \up{.036} \\

%\addlinespace[2pt]

Naive RAG
& .533 & .312 & .573 & .266
& .518 & .335 & .608 & .308 \\

+ Query Rewriting
& .555 \up{.022}
& .346 \up{.034}
& .552 \dn{.021}
& .281 \up{.015}
& .536 \up{.018}
& .368 \up{.033}
& .608 \up{.000}
& .324 \up{.016} \\

\quad $\hookrightarrow$ w/ Response CoT
& --- & --- & --- & ---
& .538 \up{.020}
& .376 \up{.041}
& .480 \dn{.128}
& .328 \up{.020} \\

\midrule
\multicolumn{9}{l}{\textit{Memory systems (raw turns + structure)}} \\
A-MEM~\shortcite{xuAMEMAgenticMemory2025}
& .531 & .308 & .571 & .267
& .515 & .326 & .609 & .307 \\

+ Query Rewriting
& .551 \up{.020}
& .344 \up{.036}
& .548 \dn{.023}
& .282 \up{.015}
% RQ
& .533 \up{.018}
& .376 \up{.050}
& .606 \dn{.003}
& .321 \up{.014} \\

\quad $\hookrightarrow$ w/ Response CoT
& --- & --- & --- & ---
& .530 \up{.015}
& .370 \up{.044}
& .470 \dn{.139}
& .331 \up{.024} \\

\addlinespace[2pt]

AnchorMem~\shortcite{shenAnchorMemAnchoredFacts2026}
& .659 & .368 & .639 & .279
& .598 & .364 & .653 & .310 \\

+ Query Rewriting
& \textbf{.754} \up{.095}
& \textbf{.524} \up{.156}
& \textbf{.732} \up{.093}
& \textbf{.432} \up{.153}
% RQ
& \textbf{.669} \up{.071}
& \textbf{.470} \up{.106}
& \textbf{.669} \up{.016}
& \textbf{.413} \up{.103} \\

\quad $\hookrightarrow$ w/ Response CoT
& --- & --- & --- & ---
& .620 \up{.022}
& .389 \up{.025}
& .523 \dn{.130}
& .314 \up{.004} \\

\midrule
\multicolumn{9}{l}{\textit{Abstractive memory system}} \\
mem0~\shortcite{chhikaraMem0BuildingProductionReady2025}
& .547 & .456 & .512 & .374
& .366 & .330 & .548& .290 \\

+ Query Rewriting
& .479 \dn{.068}
& .453 \dn{.003}
& .532 \up{.020}
& .364 \dn{.010}
% RQ
& .337 \dn{.029}
& .337 \up{.007}
& .558 \up{.010}
& .284 \dn{.006} \\
\quad $\hookrightarrow$ w/ Response CoT
& --- & --- & --- & ---
& .393 \up{.027}
& .368 \up{.038}
& .387 \dn{.161}
& .311 \up{.021} \\

Memora~\shortcite{xia2026memora}
& .608 & .445 & .600 & .387
& .501 & .388 & .617 & .333 \\
+ Query Rewriting
%wait for change
& .616 \up{.008}
& .458 \up{.013}
& .583 \dn{.017}
& .406 \up{.019}
% RQ
& .513 \up{.012}
& .403 \up{.015}
& .606 \dn{.011}
& .340 \up{.007} \\
\quad $\hookrightarrow$ w/ Response CoT
& --- & --- & --- & ---
%wait for change
& .538 \up{.037} 
& .422 \up{.034} 
& .462 \dn{.155} 
& .354 \up{.021} \\
\bottomrule
\end{tabular}
\end{adjustbox}

\caption{Joint retrieval and response evaluation across conversational query styles. \textbf{Left}: retrieval recall@10 against gold evidence turns. \textbf{Right}: end-to-end response quality. \emph{+ Query Rewriting} applies multi-facet query rewriting with RRF fusion, while \emph{w/ Response CoT} adds explicit memory selection before response generation using the same retrieved memories. Bold indicates the best-performing real memory system in each column (excluding No Memory and Oracle).}

\label{tab:joint_results}
\end{table*}

\section{Results}
\subsection{Retrieval Recall}
Table~\ref{tab:joint_results} (left) reports retrieval recall@10 across all conversational query styles. Retrieval performance drops substantially on \emph{implicit} queries compared to direct \emph{dialog} queries, confirming that conversational framing without explicit questions is significantly more challenging for current memory systems.

\textbf{AnchorMem performs the best on factual retrieval-oriented styles}, achieving the highest recall on \emph{dialog} (0.659) and \emph{counterfactual} (0.639) queries. This behavior aligns with AnchorMem's graph-structured design, which is optimized for retrieving specific factual anchors from prior interactions~\citep{shenAnchorMemAnchoredFacts2026}. In contrast, \textbf{abstractive systems (mem0 and Memora) perform best on semantically broad conversational styles}, outperforming AnchorMem by notable margins on \emph{implicit} (0.456/0.445 vs.\ 0.368) and \emph{composed} (0.374/0.387 vs.\ 0.279) queries. The abstractive memory representation appears more robust when the conversational surface-form diverges substantially from the original dialogue evidence. Overall, \emph{implicit} and \emph{composed} queries remain the hardest styles across every system, confirming that \textbf{conversational framing without an explicit question poses a fundamental retrieval challenge.}

\subsection{Response Quality}
Table~\ref{tab:joint_results} (right) presents end-to-end response quality. The ranking differs substantially from retrieval recall, and the two abstractive systems show contrasting behavior. AnchorMem achieves the strongest response quality on \emph{dialog} and \emph{counterfactual} queries, and Memora on \emph{implicit} and \emph{composed} queries, despite AnchorMem trailing both abstractive systems on \emph{implicit} and \emph{composed} retrieval. mem0, by contrast, excels at retrieval yet performs noticeably worse on downstream generation, particularly on \emph{dialog} and \emph{counterfactual} queries. This discrepancy reveals a clear \textbf{retrieval-to-response gap: semantically relevant memories do not necessarily translate into grounded conversational responses.} Crucially, the gap is not a property of abstractive memory per se: Memora, which also stores LLM-extracted memories, converts its retrieval advantage into the best implicit and composed responses, whereas mem0---strong on retrieval yet among the weakest on response quality---does not. Even the best systems remain far below the oracle ceiling on implicit and composed queries. We examine why retrieval gains fail to transfer in two places: \S\ref{sec:silent_grounding} shows part of the gap reflects a limitation of fact-recall metrics rather than the systems themselves, while \S\ref{sec:compression-vs-elaboration} traces the mem0--Memora contrast to a structural cost of lossy compression rather than abstraction itself.

\section{Analysis and Findings}
% ============================================================
% Table: Cat-5 hallucination rate, 2 answer models x thinking on/off.
%   Setup: AnchorMem top-K=10 memory shown; cat-5 questions are
%   unanswerable from the conversation. Judge: gpt-5.4-mini with v2
%   prompt (sees user name + memory block + response; flags response
%   if it asserts memory-grounded facts the memory does not support).
% Bold = better (lower-hallucination) setting per model.
% ============================================================
\newcommand{\delgood}[1]{\textcolor{green!60!black}{\scriptsize$\downarrow$#1}}

\begin{table}[t]
\centering
\footnotesize
\setlength{\tabcolsep}{4pt}
\renewcommand{\arraystretch}{1.1}
\begin{tabular}{@{}l c l l l@{}}
\toprule
\textbf{Answer model} & \textbf{Think} & \textbf{Dialog} & \textbf{Implicit} & \textbf{Mean} \\
\midrule
\multirow{2}{*}{\texttt{gemma-4-31B-it}}
  & off & 0.612 & 0.614 & 0.613 \\
  & on            & \textbf{0.504}\,\delgood{.11} & \textbf{0.594}\,\delgood{.02} & \textbf{0.549}\,\delgood{.06} \\
\midrule
\multirow{2}{*}{\texttt{qwen3.6-35B-A3B}}
  & off           & 0.592 & 0.756 & 0.674 \\
  & on  & \textbf{0.545}\,\delgood{.05} & \textbf{0.458}\,\delgood{.30} & \textbf{0.502}\,\delgood{.17} \\
\bottomrule
\end{tabular}
\caption{Hallucination rate on unanswerable conversational queries.}
\label{tab:cat5_halluc_4way}
\end{table}

\subsection{Hallucination under Conversational Framing}

We evaluate hallucination on unanswerable conversational queries, where the queried fact is absent or incorrectly attributed. For each query, the answer model receives AnchorMem's top-10 retrieved memories, and hallucination is defined as asserting unsupported memory-grounded facts.

Table~\ref{tab:cat5_halluc_4way} compares \texttt{gemma-4-31B-it} and \texttt{qwen3.6-35B-A3B} with thinking enabled and disabled.\footnote{Both models expose an \texttt{enable\_thinking} flag.} Implicit framing substantially increases hallucination when reasoning is disabled, particularly for Qwen. Enabling reasoning reduces hallucination differently across model families, but hallucination rates remain high across all settings, 
\textbf{suggesting that conversational memory hallucination remains challenging even with strong retrieval.}
% Dialog vs implicit cat-5 halluc 1.5-1.9x.
% Mis-attribution trap (Caroline / Melanie example).

\subsection{Multi-Facet Query Rewriting Narrows the Retrieval Gap}
\label{sec:rewrite}

We hypothesize that the retrieval gap on conversational queries arises from shallow surface-form matching. Conversational utterances often contain multiple latent retrieval targets, while standard retrieval typically focuses on only a single semantic aspect. For example, ``organizing my health journal from last year'' may implicitly relate to medical visits, emotional states, or wellness milestones, yet dense retrieval often retrieves memories associated with only one facet.

\paragraph{Setup.}
To address this issue, we introduce \textbf{multi-facet query rewriting}. We use \texttt{gpt-5.4-mini} to decompose each conversational query into 3--5 complementary facets spanning different entities, themes, time periods, or semantic angles. Each rewritten facet is retrieved independently, and the final ranking is aggregated using Reciprocal Rank Fusion (RRF).

\paragraph{Results.}
Table~\ref{tab:joint_results} shows that multi-facet rewriting consistently improves retrieval recall across most systems and query styles. The gains are largest for AnchorMem, with improvements of +9.5pt on dialog, +15.6pt on implicit, +9.3pt on counterfactual, and +14.7pt on composed queries. These results \textbf{suggest that conversational retrieval failures often stem from insufficient query diversification rather than purely weak memory representations.}

% However, retrieval improvements only partially translate into downstream response improvements. For example, AnchorMem’s +15.6pt gain on implicit retrieval yields only a +10.6pt improvement in response quality, indicating a remaining \emph{generation-stage bottleneck} where retrieved evidence is not fully utilized by the answer model.

% Interestingly, abstractive memory behaves differently from raw-turn memory systems. mem0 is the only system where rewriting degrades dialog performance, reducing both retrieval and response quality. Because mem0 compresses multiple mentions into canonical abstractive facts, different rewritten facets may retrieve different summaries rather than reinforcing the same underlying evidence. This \textbf{suggests that abstractive memory compression, while beneficial for semantic matching, may reduce robustness under diversified conversational retrieval.}
Interestingly, abstractive memory behaves differently from raw-turn memory systems. mem0 is
the only system where rewriting degrades \emph{dialog} performance, reducing both retrieval and
response quality, and Memora, the other abstractive system, is not degraded but gains at most
$\pm$0.02 on any style---an order of magnitude less than AnchorMem. Because both systems
canonicalize repeated mentions into a single stored statement, the different facets of a
rewritten query no longer reinforce the same underlying evidence: for mem0 they scatter across
separate summaries, and for Memora they simply re-match the one canonical index entry.
\textbf{This suggests that abstractive memory, while beneficial for semantic matching, gains
little from diversified conversational retrieval, because the surface-form redundancy that
multi-facet rewriting exploits is removed at construction time.} We return to this contrast in
\S\ref{sec:compression-vs-elaboration}.

% ============================================================
% Silent-grounding pairwise: oracle vs no-memory on the implicit
%   sub-pool where oracle's partial fact_used = 0.0 (n=332).
% Judge sees raw gold evidence turns (not the one-line gold answer),
% with A/B order randomized per case. SG-rate = (wins + 0.5*ties)/n.
% ============================================================
% ============================================================
% Silent-grounding pairwise: oracle vs no-memory on the 332 implicit
% cases where oracle's partial fact_used = 0.0. Judge sees raw gold
% evidence turns, A/B order randomized. SG = (wins + 0.5*ties)/n.
% ============================================================
% \begin{table}[t]
% \centering
% \footnotesize
% \setlength{\tabcolsep}{3pt}
% \renewcommand{\arraystretch}{1.05}
% \begin{tabular}{@{}l c c c c@{}}
% \toprule
% \textbf{Judge} & \textbf{Oracle} & \textbf{No-mem} & \textbf{Tie} \\
% \midrule
% GPT-5.4-mini    & \textbf{72.9\%} & 26.8\% & 0.3\%\\
% Opus-4.7        & \textbf{75.3\%} & 14.8\% & 9.9\%\\
% \bottomrule
% \end{tabular}
% \caption{Silent-grounding pairwise judgment where oracle's partial \texttt{fact\_used} = 0.0.}
% \label{tab:silent_grounding}
% \end{table}
\begin{table}[t]
\centering
\small
\setlength{\tabcolsep}{5pt}
\renewcommand{\arraystretch}{1.05}
\begin{tabular}{@{}lccc@{}}
\toprule
\textbf{Comparison} & \textbf{\emph{Faithfulness}} & \textbf{\emph{Relevance}} & \textbf{\emph{Engagement}} \\
\midrule
Oracle vs no-mem & \textbf{+55.1} & +18.1 & +31.0 \\
Oracle vs random & \textbf{+33.7} & +9.3 & +22.3 \\
\bottomrule
\end{tabular}
\caption{Pairwise preference margin on the subset of implicit queries exhibiting silent grounding}
\label{tab:silent_grounding_per_dim_winrate}
\end{table}

\subsection{Silent Grounding: Beyond Explicit Fact Recall}
\label{sec:silent_grounding}

Comparing the retrieval and response halves of
Table~\ref{tab:joint_results} reveals a consistent gap: retrieval
improvements only partially translate into response gains. For
example, AnchorMem's multi-facet rewrite improves implicit retrieval
recall by $+$15.6pt but response quality by only $+$10.6pt. While
this may suggest that answer models fail to utilize retrieved
evidence, part of the gap may instead reflect a limitation of
explicit fact-based evaluation. Our \texttt{fact\_used} metric scores
whether the response surfaces the gold fact, but on \textbf{implicit
queries memory can still improve responses without directly stating
that fact}---for example, through contextual grounding, appropriate
tone, or relevant follow-up questions. We refer to this phenomenon as
\textbf{\emph{silent grounding}.}

\paragraph{Setup.}
We analyze the 332 implicit-query cases where oracle retrieval still
receives \texttt{fact\_used}~=~0.0. We compare three response variants:
\textbf{Oracle} (gold evidence turns), \textbf{no-mem} (no memory
provided), and \textbf{random} (three random non-gold turns from the
same conversation as a memory control). For each case we score every
variant's response with Claude-Opus-4.7 on three independent criteria
(\emph{faithfulness, relevance, engagement}; each scored $0/0.5/1$), and
report pairwise dimension margins (variant-A win rate $-$ variant-B
win rate).

\paragraph{Results and Implication.}
Table~\ref{tab:silent_grounding_per_dim_winrate} shows that Oracle memory substantially outperforms both no-memory and random-memory baselines, particularly on faithfulness ($+$55.1pt vs no-mem) and engagement ($+$31.0pt). These results suggest that conversational memory often improves contextual grounding without explicitly surfacing the gold fact, implying that strict fact-recall metrics alone underestimate the value of retrieval on implicit queries. Appendix~\ref{qualitative-analysis} provides qualitative examples comparing the three settings.
\begin{table}[t]
\centering
\small
\setlength{\tabcolsep}{6pt} % 稍微調寬字元間距，讓四欄比例更美觀
\renewcommand{\arraystretch}{1.05}
\begin{tabular}{@{}lccc@{}}
\toprule
\textbf{Method} & \textbf{\emph{Faithfulness}} & \textbf{\emph{Relevance}} & \textbf{\emph{Engagement}} \\
\midrule
Naive RAG  & $+1.1$ & $+4.1$ & \textbf{+37.6} \\
A-MEM     & $+0.8$ & $+4.3$ & \textbf{+37.6} \\
AnchorMem & $+3.0$ & $+4.7$ & \textbf{+39.7} \\
mem0      & $+2.3$ & $+0.7$ & \textbf{+30.9} \\
Memora    & $+2.4$ & $+1.4$ & \textbf{+31.2} \\
\bottomrule
\end{tabular}
\caption{Pairwise preference margin (CoT win rate $-$ Oracle win rate) on implicit queries.}
\label{tab:e3_per_dim_winrate}
\end{table}

\iffalse
\begin{table}[t]
\centering
\small
\setlength{\tabcolsep}{3.5pt}
\renewcommand{\arraystretch}{1.05}
\begin{tabular}{@{}lcccc@{}}
\toprule
\textbf{Dimension} & \textbf{NaiveRAG} & \textbf{A-MEM} & \textbf{AnchorMem} & \textbf{mem0} \\
\midrule
\emph{Faithfulness} & $+1.1$ & $+0.8$ & $+3.0$ & $+2.3$ \\
\emph{Relevance}    & $+4.1$ & $+4.3$ & $+4.7$ & $+0.7$ \\
\emph{Engagement}   & \textbf{+37.6} & \textbf{+37.6} & \textbf{+39.7} & \textbf{+30.9} \\
\bottomrule
\end{tabular}
\caption{Pairwise preference margin (CoT win rate $-$ Oracle win rate) on implicit queries. CoT-selected responses substantially improve engagement, while faithfulness and relevance remain largely unchanged.}
\label{tab:e3_per_dim_winrate}
\end{table}

\begin{table}[t]
\centering
\small
\setlength{\tabcolsep}{3.5pt}
\renewcommand{\arraystretch}{1.05}
\begin{tabular}{@{}lccccc@{}}
\toprule
\textbf{Dimension} & \textbf{BM25} & \textbf{Dense} & \textbf{A-MEM} & \textbf{AnchorMem} & \textbf{mem0} \\
\midrule
\emph{Faithfulness} & $-6.6$ & $+1.1$ & $+0.8$ & $+3.0$ & $+2.3$ \\
\emph{Relevance}    & $+1.1$ & $+4.1$ & $+4.3$ & $+4.7$ & $+0.7$ \\
\emph{Engagement}   & \textbf{+33.1} & \textbf{+37.6} & \textbf{+37.6} & \textbf{+39.7} & \textbf{+30.9} \\
\bottomrule
\end{tabular}
\caption{Pairwise preference margin (CoT win rate $-$ Oracle win rate) on implicit queries. CoT-selected responses substantially improve engagement, while faithfulness and relevance remain largely unchanged.}
\label{tab:e3_per_dim_winrate}
\end{table}
\fi
\subsection{Chain-of-Thought Selection Approaches Oracle Quality}
\label{sec:cot}

The default setting directly feeds all retrieved top-$K$ memories into the answer model. We compare this against a \emph{CoT} variant, where the model first explicitly selects relevant memories before generating its response. The retrieved top-$K$ remains identical; only the answer-side prompt changes.

\paragraph{Mixed effect on response quality.}
As shown in Table~\ref{tab:joint_results}, \emph{CoT} consistently improves dialog, \emph{implicit}, and \emph{composed} responses, but substantially hurts \emph{counterfactual} performance across all systems. On counterfactual queries, \emph{CoT} prompting lowers performance because the model first restates the user's message before consulting memory: the false premise becomes the framing of the response, and the retrieved memory is then reconciled with it rather than used to correct it. Asking the model to explicitly select the memories it uses recovers only a small part of this loss (Appendix~\ref{app:cot_ablation}).
% We find that explicit memory selection biases the model toward reconciling retrieved memories with the user’s framing, increasing unaware or sycophantic responses instead of correcting false premises.

\paragraph{CoT-selected responses approach oracle quality.}
% We compare \emph{CoT} responses against Oracle responses on the full implicit set using Claude-Opus-4.7 pairwise evaluation over faithfulness, relevance, and engagement. As shown in Table~\ref{tab:e3_cot_vs_oracle}, \emph{+cot} responses are preferred in 41--47\% of cases, while Oracle wins only 11--21\%. The gains are primarily driven by stronger conversational engagement, suggesting that broader retrieved context can sometimes support more grounded responses than the minimal gold evidence alone.+cot
We compare \emph{CoT} responses against Oracle on the full implicit
set using the same Claude-Opus-4.7 pairwise evaluation as \S~\ref{sec:silent_grounding}. Table~\ref{tab:e3_per_dim_winrate} shows that there is \textbf{a huge gap when it comes to response engagement}: \emph{CoT} beats Oracle by
$+$31 to $+$40 points on engagement, while faithfulness and relevance
margins stay within $\pm$7 points across systems. Broader retrieved
context can therefore support more grounded conversational responses
without sacrificing factual reliability.

\paragraph{Augmenting \textsc{LoCoMo-Conv} with supportive memory.}
Motivated by this observation, we extend \textsc{LoCoMo-Conv} with a \texttt{supportive\_memory} field for \emph{implicit} queries, containing conversational turns frequently selected by successful \emph{CoT} responses. The original \texttt{evidence} annotations remain unchanged, while \texttt{supportive\_memory} provides an auxiliary conversational-support context for future evaluation. For detailed qualitative analysis, please refer to Appendix~\ref{qualitative-analysis}.

\subsection{Compression versus Elaboration in Memory Construction}
\label{sec:compression-vs-elaboration}

The rewriting results in \S\ref{sec:rewrite} suggest an implication that extends beyond
query rewriting itself. Multi-facet rewriting yields the largest improvement on implicit
retrieval (+15.6 for AnchorMem) by expanding an underspecified conversational utterance into
multiple semantically explicit facets before retrieval. Rather than introducing new evidence,
the rewriting process exposes semantic aspects already implied by the user's utterance,
allowing them to align more readily with the relevant memory. This observation suggests that
the primary challenge of implicit conversational retrieval is not the absence of evidence,
but semantic underspecification: conversational surface forms often fail to express the
concepts necessary for successful memory matching.

This naturally raises a broader question: \textit{if semantic elaboration improves retrieval when
applied at query time, can the same principle be incorporated during memory construction?}
The two abstractive memory systems provide evidence in favor of this hypothesis. Both process
interactions before storage and achieve the strongest retrieval performance on the most
challenging query styles, reaching implicit/composed recall of 0.456/0.374 for mem0 and
0.445/0.387 for Memora, compared with 0.368/0.279 for AnchorMem. These results suggest that
enriching memory representations prior to storage, rather than preserving raw dialogue turns
alone, substantially improves conversational retrieval.

Retrieval performance alone, however, does not guarantee better responses. The way in which
memory is abstracted determines whether retrieved information remains useful for response
generation. Here the two systems diverge. mem0 constructs memory primarily through
compression, merging repeated observations into concise summary facts. Although this
representation appears sufficient for semantic matching, it yields one of the weakest
response-generation results (dialog \texttt{fact\_used}: 0.366 versus 0.598 for AnchorMem),
and its strong implicit retrieval does not translate into corresponding gains in implicit
response quality. Memora instead adopts a different abstraction strategy: retrieval operates
over a compact abstractive index augmented with cue anchors, while the retrieved memory
retains detailed original content for generation. Despite achieving retrieval performance
comparable to mem0, Memora attains the best implicit \texttt{fact\_used} (0.388) and composed
coverage (0.333). These results suggest that abstraction itself is not detrimental. Rather,
the limitation arises when abstraction becomes \emph{lossy}: compression preserves sufficient
semantic information for retrieval while discarding the concrete details required to ground
a response.

The rewriting experiments further clarify the relationship between query-time and
memory-time elaboration. If semantic expansion has already been incorporated into stored
memory representations, additional elaboration at retrieval time should offer only marginal
benefit. Empirically, this is exactly what we observe. Multi-facet rewriting improves
AnchorMem's retrieval by +0.09 to +0.16 recall, yet changes Memora by at most $\pm0.02$ and
mem0 by only $-0.07$ to $+0.02$. Query rewriting and memory elaboration therefore appear to
serve largely overlapping roles, suggesting that they address the same underlying source of
retrieval failure---semantic underspecification---at different stages of the memory pipeline.

Taken together, these observations point toward a broader design principle for conversational
memory systems. Memory construction should increase semantic accessibility beyond raw
dialogue turns while simultaneously preserving the specific information required for response
grounding. In other words, this suggests that future conversational memory systems may benefit more from semantic elaboration than lossy compression. We emphasize, however, that this comparison should not be interpreted as a
definitive evaluation of compression versus elaboration strategies. The two abstractive
systems differ in several design choices beyond their memory-construction mechanisms (e.g.,
Memora's cue-anchor design), and these factors may also contribute to the observed
differences. A controlled comparison that isolates memory-construction strategies from other
architectural factors remains an important direction for future work.
\section{Conclusion}
We introduced \textsc{LoCoMo-Conv}, a benchmark for evaluating whether
memory-augmented agents can invoke memory under realistic conversational
framing rather than explicit QA. Our findings are fourfold. First,
conversational framing reveals retrieval and response gaps hidden by QA-style
evaluation, especially for implicit and composed queries. Second, strict
fact-recall metrics miss the \emph{silent grounding} we observe on implicit
queries, where memory improves responses without explicitly surfacing the gold
fact. Third, while both multi-facet query rewriting and abstractive memory
improve retrieval (specifically \emph{implicit} and \emph{composed} style), abstractive compression often removes details needed for
grounded responses, suggesting that reasoning-based memory
\emph{elaboration} is more promising than lossy compression. Finally, we
release \texttt{supportive\_memory}, an auxiliary annotation layer capturing
conversationally supportive context beyond the original gold evidence.

\section*{Limitations and Future Work}
\textsc{LoCoMo-Conv} has several limitations. Conversational rewrites and
\texttt{supportive\_memory} annotations are generated through LLM-based pipelines and may
inherit model-specific biases; human validation (Appendix~\ref{app:human_annotation}) covers a
40-item sample per style rather than the full set. Our evaluation relies on a single
open-weights answer model and one primary judge family, with cross-judge and human validation
performed on subsets. The benchmark is built on the ten LoCoMo conversations, which is small
relative to real-world long-horizon interactions; however, the construction pipeline is
source-agnostic---it takes any conversation with QA-style evidence annotations and applies the
same rewriting and clustering procedure---so the benchmark can be scaled to larger or newer
conversation pools without changing the methodology. Future work could extend evaluation to
broader model families and to such larger pools, and improve the reliability of
supportive-memory annotations.
% \textsc{LoCoMo-Conv} has several limitations. Conversational rewrites and
% \texttt{supportive\_memory} annotations are generated through LLM-based
% pipelines and may inherit model-specific biases. Our evaluations mainly rely on
% a single answer-model and judge-model family, with cross-model validation
% performed only on subsets. In addition, the \emph{CoT} setting jointly changes
% memory selection and prompting strategy, leaving their individual effects
% unisolated. The underlying conversation pool also remains relatively small
% compared to real-world long-horizon interactions. Finally, our main results
% (Table~\ref{tab:joint_results}) are reported as single-run scores under
% deterministic decoding (temperature$=0$); we do not report variance across
% random seeds or judge samples, so small between-system differences should be
% interpreted with caution. Future work could extend evaluation to broader model
% families, longer conversations, multiple runs with variance reporting, human
% validation, and more reliable supportive-memory annotations.

\section*{Acknowledgments}
This work was financially supported by the National Science and Technology Council (NSTC) and the Featured Area Research Center Program within the framework of the Higher Education Sprout Project by the Ministry of Education in Taiwan, under Grants 112-2223-E-002-012-MY5, 115-2628-E-002-023-MY4, and 115L900901. We also thank Chia-En Hsu and Chih-Chih Yang for
their help with data annotation. We used AI assistants to support manuscript editing, language refinement, and presentation. All research design, experiments, analyses, and conclusions were developed and verified by the authors.

\bibliography{custom}
% ============================================================
%                       APPENDIX
% ============================================================
\appendix

\section{LLM Judge Validation}
As shown in Table \ref{tab:judge_kappa}.
\label{app:judge_validation}
% Cross-judge agreement: response judge run on 1488-item subset with 3 LLMs
% (gpt-5.4-mini, claude-sonnet-4.5, qwen3.6-35b-a3b, all with reasoning/thinking enabled).
% Report Cohen's pairwise kappa and Fleiss' 3-way kappa.
% Per-style breakdown.

\begin{table}[h]
\centering\small
\begin{tabular}{@{}lcccc@{}}
\toprule
Models & Cohen $\kappa$ & Strength \\
\midrule
gpt-5.4-mini vs claude-sonnet-4.5 & 0.761 & substantial \\
gpt-5.4-mini vs qwen3.6-35b      & 0.681 & substantial \\
claude-sonnet-4.5 vs qwen3.6-35b & 0.797 & substantial \\
\midrule
\textbf{Fleiss 3-way $\kappa$}    & \textbf{0.745} & \textbf{substantial} \\
\bottomrule
\end{tabular}
\caption{Cross-judge agreement on response judging}
\label{tab:judge_kappa}
\end{table}
\section{Data Statistics}
\label{app:data_stats}

\textsc{LoCoMo-Conv} attaches conversational rewrites to each QA item
in LoCoMo10. Table~\ref{tab:data_qa_stats} lists the per-style
counts. Dialog and implicit share the full 1{,}986-item QA pool;
counterfactual excludes the 446 cat-5 adversarial items whose premise
has no gold answer; composed clusters are constructed by combining
two source QAs whose gold evidence overlaps.

\begin{table}[h]
\centering
\footnotesize
\setlength{\tabcolsep}{8pt}
\renewcommand{\arraystretch}{1.05}
\begin{tabular}{@{}lr@{}}
\toprule
\textbf{Style} & \textbf{Items} \\
\midrule
Dialog          & 1{,}986 \\
Implicit        & 1{,}986 \\
Counterfactual  & 1{,}540 \\
Composed clusters &  1{,}069\\
\bottomrule
\end{tabular}
\caption{Per-style query counts in \textsc{LoCoMo-Conv}.}
\label{tab:data_qa_stats}
\end{table}

% \section{Per-Category Breakdown}
% \label{app:per_cat}
% Full per-cat table by (style, system, variant) for top_k K=10 on FULL 5812.
% Show cat 1 (list), cat 2 (date), cat 3 (multi-hop), cat 4 (open commonsense), cat 5 (halluc) separately.

% ============================================================
% Appendix: Prompts used in the pipeline
% ============================================================
\section{Prompts}
\label{app:prompts}
All four conversational query styles in \textsc{LoCoMo-Conv} are
generated by prompting \texttt{gpt-5.4-mini} to rewrite original LoCoMo10
content into a first-person utterance. For the three single-QA
styles (Dialog, Implicit, Counterfactual), each API call uses a
\emph{system} message (the style-specific instruction below) plus a
shared \emph{user} message template carrying the QA data. The
Composed style takes multiple source QAs as input and is therefore
delivered as a single user message that interpolates both the task
instruction and the member memories.

\subsection{Conversational query rewrite prompts}
\label{app:rewrite_prompts}

\paragraph{Shared user prompt template}
\begin{lstlisting}[breaklines=true,basicstyle=\small\ttfamily]
speaker_a: {speaker_a}
speaker_b: {speaker_b}

Category: {category} ({category_desc})
Question: {question}
Gold answer: {answer}

Rewrite this as one first-person utterance from the subject speaker.
\end{lstlisting}

\subsubsection{Dialog rewrite}
\begin{lstlisting}[breaklines=true,basicstyle=\small\ttfamily]
You convert third-person QA pairs into natural first-person dialog turns (EXPLICIT memory queries).

Setting: The two people in `conversation` (speaker_a, speaker_b) have been chatting with an AI assistant for a long time. The assistant has memories of everything they've shared. Now one of them opens a fresh chat with the assistant and says ONE message that should naturally make the assistant look up the right memory and answer the original question.

Your job, for one QA at a time:
1. Choose the subject speaker -- the participant the question is asking ABOUT (e.g. "What did Caroline research?" -> Caroline). The dialog turn must come from THAT speaker, addressed to the assistant in first person ("I", "me", "my").
2. Write a single natural utterance that this speaker would actually send as a DIRECT memory query (e.g. "Hey, do you remember when I painted that sunrise?", "Remind me what I was researching last spring?"). The utterance must NOT contain the answer, dates, or evidence specifics.
3. Rate naturalness:
   - "high"   = sounds like a real thing someone would say in chat
   - "medium" = slightly contrived but plausible
   - "low"    = a real person would NOT directly ask this. The information is the kind that comes up implicitly through context (advice-seeking, sharing a feeling, describing a situation), not through a direct "do you remember X" question.
4. Reason field: one short sentence justifying naturalness; for "low", also hint at what implicit context would naturally surface this memory.

Return STRICT JSON with keys: subject_speaker, dialog_query, naturalness, reason. No preamble, no code fences.
\end{lstlisting}

\subsubsection{Implicit rewrite}
\begin{lstlisting}[breaklines=true,basicstyle=\small\ttfamily]
You convert third-person QA pairs into IMPLICIT first-person dialog turns.

Setting: The two people in `conversation` (speaker_a, speaker_b) have been chatting with an AI assistant for a long time. The assistant has memory of everything they've shared. Now the subject speaker opens a fresh chat and says ONE message. The message presents a real-life CONTEXT (a feeling, situation, problem, plan, decision they're facing) where the gold memory below would be the relevant thing for the assistant to recall, surface, or apply on its own initiative.

Hard rules:
- The utterance must NOT directly ask "do you remember...", "remind me...", "what did I say about...". It must NOT mention the answer or any evidence detail.
- The utterance should sound like the start of a normal conversation -- a vent, a plan, a decision, a question about life -- that a memory-aware assistant could respond to better by drawing on the gold memory.
- The expected_memory_use field describes what an ideal assistant response would look like: which memory it should surface, and how it would apply it. Be concrete.

Examples:
- Memory: "Melanie does running, reading, violin for self-care"
  implicit_query: "Ugh, I've been so stressed this week, I literally can't unwind."
  expected_memory_use: "Suggest the user try the self-care routines she's mentioned before -- going for a run, reading, or playing violin -- instead of giving generic stress tips."

- Memory: "Caroline is researching adoption agencies for the summer"
  implicit_query: "Trying to plan out what I'm doing the next few months and I feel kind of stuck."
  expected_memory_use: "Bring up the adoption-agency research she said was her summer focus, and help her break it into steps."

Return STRICT JSON with keys: subject_speaker, implicit_query, expected_memory_use, reason. No preamble, no code fences.
\end{lstlisting}

\subsubsection{Counterfactual rewrite}
\begin{lstlisting}[breaklines=true,basicstyle=\small\ttfamily]
You convert third-person QA pairs into COUNTERFACTUAL first-person dialog turns.

Setting: The subject speaker is talking to an AI assistant that has memory of past conversations. The speaker now asserts a FACTUALLY WRONG version of something they previously shared, and either asks for confirmation or states it casually in passing. An ideal assistant would catch the inconsistency with stored memory and gently push back.

Hard rules:
1. Choose the subject speaker -- the participant the original question is asking ABOUT. The utterance must come from THAT speaker, in first person.
2. Generate a PLAUSIBLE-WRONG counterfactual:
   - Cat 1 (single-hop fact): replace the gold fact with a plausible alternative of the same type.
   - Cat 2 (temporal): shift the date by 2-5 years or change the month/season noticeably.
   - Cat 3 (inference): assert the OPPOSITE inference.
   - Cat 4 (multi-hop fact): substitute one element of the chain with a plausible wrong.
3. Preserve enough CONTEXT so the assistant can reason: keep the topic anchor named, keep a "now" anchor if the original was temporal, keep first-person pronouns.
4. Phrase the wrong assertion in a casual, conversational way -- NOT a quiz.
5. Do NOT mention the gold (correct) answer in the utterance.

Return STRICT JSON with keys: subject_speaker, counterfactual_query, asserted_wrong, reason. No preamble, no code fences.
\end{lstlisting}

\subsubsection{Composed rewrite}
\begin{lstlisting}[breaklines=true,basicstyle=\small\ttfamily]
You compose a single first-person utterance that requires MULTIPLE memories from a user's past conversations to answer well.

Setting: The subject speaker is opening a fresh chat with an AI assistant that has memory of all past conversations. They send ONE message that should make the assistant draw on EVERY memory listed below to construct a good response.

Hard rules:
- The utterance must come from one speaker, in first person.
- The utterance must NOT directly name the gold answers or quote evidence text.
- It should sound natural -- a real situation, plan, decision, or reflection where ALL of the listed memories are relevant.
- It should NOT be a generic question that any memory could satisfy -- only the listed memories together should fully address it.

Member memories (each is a Q+A from a past evaluation; the assistant should "use" the answer when responding):
{members_block}

speaker_a: {speaker_a}
speaker_b: {speaker_b}

Return STRICT JSON with keys: subject_speaker, composed_query, expected_memory_use, reason. No preamble, no code fences.
\end{lstlisting}

\noindent

\subsection{Answer prompt}
\begin{lstlisting}[breaklines=true,basicstyle=\small\ttfamily]
You are an AI assistant with long-term memory of past conversations with the user.

Below are relevant memory items the assistant has access to:
<memory>
{memory_block}
</memory>

The user (speaker: {speaker}) now says:
"{query}"

Provide a concise answer that directly addresses the user. If the memory clearly contains the relevant information, use it. If the memory does not contain the needed information, say so plainly.

Answer:
\end{lstlisting}

\subsection{No-memory baseline}
\begin{lstlisting}[breaklines=true,basicstyle=\small\ttfamily]
You are an AI assistant. You do not have access to any prior conversation history with this user.

The user (speaker: {speaker}) says:
"{query}"

Provide a concise answer that directly addresses the user. If you do not have the information needed to answer specifically, say so plainly.

Answer:
\end{lstlisting}

\subsection{Multi-facet query rewrite}
\begin{lstlisting}[breaklines=true,basicstyle=\small\ttfamily]
A user has said the following in conversation with an AI assistant. The AI needs to retrieve relevant memories from past conversations to respond well.

Generate 3 to 5 SHORT search queries, each capturing a DIFFERENT facet of what would be relevant: different entities, time periods, themes, or related angles. Each query should be self-contained and search a distinct angle (don't paraphrase the same thing).

User message:
"{query}"

Output ONLY the queries, one per line, no bullets, no numbering, no preamble. Each under 15 words.
\end{lstlisting}

\subsection{Partial-credit \texttt{fact\_used} judge (dialog and implicit)}
\begin{lstlisting}[breaklines=true,basicstyle=\small\ttfamily]
You are evaluating whether an AI assistant's response correctly conveys a target fact from the user's past conversation memory.

User (named {user_name}): {query}

Target fact the response should convey:
{gold}

Assistant's response:
{response}

Score the response on a 3-level scale:

- 1.0 (full): the substance of the target fact is correctly conveyed. Paraphrasing is fine; verbatim not required. Hedging or qualifying language ("I think...", "if I recall...", "based on what we discussed...") is fine. Minor stylistic differences are fine.

- 0.5 (partial): the response captures the central concept or category of the gold fact but misses or only vaguely references the specific items / sub-facts the gold lists. Examples: the gold names three items but the response names two of them or refers to them only as a category ("pastries" instead of "croissants, muffins, and tarts"); the gold lists two emotions but the response covers one.

- 0.0 (none): the response asserts incorrect content that conflicts with the gold, gives only a vague allusion with no specific substance, or completely omits the fact.

Output ONLY one number: 1.0, 0.5, or 0.0.
\end{lstlisting}

\subsection{Counterfactual 3-way judge}
\begin{lstlisting}[breaklines=true,basicstyle=\small\ttfamily]
You are evaluating an AI assistant's response to a user message that contains a FALSE PREMISE about the user's own past. The assistant has access to the user's true memory.

User message (contains a false premise):
{query}

Ground-truth fact (what the memory actually says):
{gold}

Assistant's response:
{response}

Classify the response into exactly one category:

- A (unaware): the response shows no signal that the assistant noticed any conflict between the user's claim and memory. This includes responses that treat the user's claim as if it were true, extend or elaborate on the false premise, redirect to an unrelated topic without flagging the conflict, or give a generic reply that ignores the mismatch entirely.

- B (aware, no correction): the response shows some awareness that the user's claim does not match memory -- by hedging ("I don't have a record of that...", "I'm not sure that's right..."), deflecting, refusing to confirm, or asking the user to clarify -- but it does NOT state the ground-truth fact. The user is left without the correct information.

- C (corrected): the response states the ground-truth fact (or a clear paraphrase of it), regardless of whether it explicitly points out the user's error. Simply providing the correct information counts as C, even if the response does not call out the user's mistake. Hedging is fine as long as the truth is conveyed (e.g., "I think it was actually...").

Output ONLY one letter: A, B, or C.
\end{lstlisting}

\subsection{Composed atomic-fact coverage judge}
\begin{lstlisting}[breaklines=true,basicstyle=\small\ttfamily]
You are evaluating whether an AI assistant's response covers each of several atomic facts.

User's message:
{query}

Atomic gold facts the response should cover (one per line, numbered):
{facts_block}

Assistant's response:
{response}

For EACH atomic fact, decide whether the response covers it (explicitly or by clear semantic equivalence -- paraphrasing is OK, but the substance must be present). Implicit/vague mentions that a reader couldn't reasonably extract count as NOT covered.

Output STRICT JSON ONLY:
{
  "judgments": [
    {"fact": 1, "covered": true, "evidence": "<short quote from response>"},
    {"fact": 2, "covered": false, "evidence": "<why not>"},
    ...
  ]
}
One entry per fact. No preamble, no markdown.
\end{lstlisting}

\subsection{Hallucination judge}
\begin{lstlisting}[breaklines=true,basicstyle=\small\ttfamily]
You are evaluating whether an AI assistant hallucinated a memory-grounded answer when the retrieved memory does not support one.

User (named {user_name}): {query}

Retrieved memory shown to the assistant:
{memory_block}

Assistant's response:
{response}

The retrieved memory does NOT contain the information needed to answer this question. Did the assistant assert specific facts as if they came from this memory (i.e., misuse the memory to fabricate a memory-grounded answer)?

- YES = response asserts memory-grounded claims that the shown memory does not support (e.g., "based on our chats you mentioned X" when no such turn is in the memory).
- NO = response declines, asks for clarification, answers from general world knowledge without claiming memory support, or only restates facts that are actually present in the shown memory.

Output ONLY one word: YES or NO.
\end{lstlisting}
\subsection{Per-dimension scoring judge}

\begin{lstlisting}[breaklines=true,basicstyle=\small\ttfamily]
You are evaluating how appropriately an AI assistant responded to a user, given the memory items it had access to.

User (named {user_name}) says:
{query}

Memory items the assistant could draw on (from prior conversations):
{memory_block}

Assistant's response:
{response}

Score the response on three independent criteria. For each, output one of: 1, 0.5, or 0.

(1) faithfulness -- whether the response is well-grounded in the memory items above:
  1   = clearly draws on a specific memory item (paraphrasing is fine)
  0.5 = consistent with memory but does not actively use any specific item (neutral coexistence)
  0   = contradicts memory OR fabricates plausible-sounding specifics not present in the memory

(2) relevance -- whether the response addresses what the user is asking about or describing:
  1   = directly addresses the user's question / situation
  0.5 = partially addresses; some of the response is on-topic and some is generic
  0   = off-topic / pure boilerplate / redirects to an unrelated subject

(3) engagement -- how the response engages with the user's emotional / situational framing:
  1   = acknowledges the user's state AND offers something concrete (a fitting follow-up, an actionable suggestion, or genuine empathy)
  0.5 = polite, functional acknowledgment -- neutral and on-topic but does not go beyond a generic "I see / I'm sorry to hear that / could you tell me more"
  0   = cold refusal, dismissive, pure list, or ignores the user's emotional/situational framing entirely

Output STRICT JSON ONLY:
{
  "faithfulness": 1 | 0.5 | 0,
  "relevance":    1 | 0.5 | 0,
  "engagement":   1 | 0.5 | 0
}
No preamble, no markdown.
\end{lstlisting}

% \subsection{Silent-grounding pairwise judge}
% \begin{lstlisting}[breaklines=true,basicstyle=\small\ttfamily]
% You are comparing two AI assistant responses to the same user message. The assistant is in a long-running conversation with this user.

% User (named {user_name}) says:
% {query}

% Relevant excerpts from the user's actual past conversations (use these to understand the user's situation):
% {evidence_block}

% ---
% Response A:
% {response_a}
% ---
% Response B:
% {response_b}
% ---

% Which response gives a more appropriate reply given the user's actual situation? Judge how well each response fits the context -- its framing, relevance, tone, and engagement with what the user is actually dealing with.

% Consider:
% - Does the response engage with the user's situation in a way that fits the context, or does it feel generic / off-topic / disconnected?
% - Does it respond to what the user is actually asking or feeling?
% - A response that confidently asserts a fact that contradicts the excerpts is worse than one that hedges honestly.

% If both are equally appropriate (or equally inappropriate), output TIE.

% Output ONLY one of: A, B, TIE.
% \end{lstlisting}

% Bibliography entries for the entire Anthology, followed by custom entries
%\bibliography{anthology,custom}
% Custom bibliography entries only

\section{Adversarial Example}
See Table~\ref{tab:adversarial}.
\begin{table*}[t]
\centering
\small
\renewcommand{\arraystretch}{1.3}
\setlength{\tabcolsep}{6pt}
\begin{tabular}{@{} >{\raggedright\arraybackslash}p{0.48\textwidth} | >{\raggedright\arraybackslash}p{0.48\textwidth} @{}}
\toprule
\multicolumn{1}{c}{\textbf{Knowledge Context}} & \multicolumn{1}{|c}{\textbf{Conversational Query Styles}} \\
\midrule

\textbf{Original Question:} \newline
\emph{What did Caroline realize after her charity race?} \vspace{0.5em} \newline
\textbf{Gold Answer:} \newline
\emph{(unanswerable)} \vspace{0.5em} \newline
\textbf{Memory:} \newline
$\bullet$ \texttt{[D2:3]} \textbf{Melanie}: ``\ldots I'm starting to realize that self-care is really important\ldots'' \newline
\footnotesize\textcolor{gray}{$\leftarrow$ The memory exists, but belongs to \emph{Melanie}, not Caroline.}
&
\textbf{Dialog Question:} \newline
\emph{Do you remember what I told you I realized after that charity race I did?} \vspace{0.8em} \newline
\textbf{Implicit User Turn:} \newline
\emph{I'm thinking about signing up for another charity run, but I'm not sure if it's actually the right way for me to give back.} \vspace{0.8em} \newline
\textbf{Counterfactual User Turn:} \newline
\emph{(not generated --- no gold fact exists to contradict)} \\

\bottomrule
\end{tabular}
\caption{Example of an unanswerable conversational query. The retrieved memory contains the adversarial claim, but it belongs to a different speaker. Hallucination occurs when the assistant incorrectly attributes this memory to the user.}
\label{tab:adversarial}
\end{table*}
\section{Qualitative Analysis}
\label{qualitative-analysis}
% =====================================================================
% Prose paragraphs to accompany the two qualitative tables:
%   - qual_random_control  (Table tab:qual_random_control)
%   - qual_cot_supportive  (Table tab:qual_cot_supportive)
%
% Drop into Section 7.3 (silent grounding) and Section 7.4 (CoT)
% respectively. Each paragraph is intended to sit immediately before
% or after its referenced table.
% =====================================================================

% ---------------------------------------------------------------------
% Paragraph A: silent-grounding qualitative analysis
% Belongs in Section 7.3 (Silent Grounding)
% ---------------------------------------------------------------------
\paragraph{Silent Grounding}
Table~\ref{tab:qual_random_control} illustrates a representative implicit query where the Oracle response receives \texttt{fact\_used}=0 despite being clearly grounded in the user’s conversational history. Although the gold fact (\emph{``writing a travel blog''}) is never explicitly surfaced, the response synthesizes related memories about writing novels and sharing stories into a personalized weekend suggestion aligned with the user’s interests. In contrast, the random-memory control falls back to a generic clarification response, suggesting that the effect arises from relevant conversational grounding rather than merely providing additional context.
\paragraph{Supportive Memory}
Drawing from an AnchorMem case in which \emph{CoT} wins the pairwise
judgment, Table~\ref{tab:qual_cot_supportive} shows how \emph{CoT}
retrieves supportive conversational context beyond Oracle's narrow
gold evidence. Oracle only receives the turn describing the user's
anxiety before the studio opening, whereas \emph{CoT} additionally
selects an earlier conversation expressing a similar emotional state
during the studio's setup phase. This broader context enables a more
emotionally grounded response that connects the user's past and
present experiences, illustrating the type of conversational support
captured by the released \texttt{supportive\_memory} annotations.

\begin{table*}[t]
\centering
\footnotesize
\setlength{\tabcolsep}{4pt}
\renewcommand{\arraystretch}{1.2}
\begin{tabular}{@{}p{0.08\textwidth} p{0.46\textwidth} p{0.40\textwidth}@{}}
\toprule
\multicolumn{3}{@{}p{0.96\textwidth}@{}}{\textbf{User query.} \emph{``I finally have some free time on the weekends, but I'm bored with my current routine and want to start something new that actually feels meaningful.''}} \\
\midrule
\textbf{Variant} & \textbf{Memory shown to the answer model} & \textbf{Response} \\
\midrule
\textbf{Oracle} &
{\small \textbf{[D4:1, 2 Aug 2023]} Tim: \emph{``Hey John! How've you been? Something awesome happened - \textbf{I'm writing articles about fantasy novels} for an online mag. It's so rewarding!''}} \newline
{\small\textbf{[D6:6, 11 Aug 2023]} Tim: \emph{``\dots \textbf{I have been writing more articles} - it lets me combine my \textbf{love for reading and the joy of sharing great stories}\dots''}} \newline
{\small\textbf{[D15:3, 21 Oct 2023]} Tim: \emph{``\dots My \textbf{writing is going well: I'm in the middle of a fantasy novel} and it's a bit nerve-wracking but so exciting\dots \textbf{Writing brings such joy}\dots''}} \newline
{\small\textbf{[D27:37, 2 Jan 2024]} Tim: \emph{``\textbf{I love traveling} too. That picture is awesome. Have you been to Paris?\dots''}}
&
\emph{``Since you've \textbf{found so much joy and meaning in writing}---from your \textbf{articles on fantasy novels to working on your own novel}---perhaps you could \textbf{dedicate your weekends to finishing your book or exploring a new writing project}?''} \\
\addlinespace[2pt]
\textbf{Random} &
{\small\textbf{[D2:6, 15 Jun 2023]} John: \emph{``Thanks! Exciting times ahead! I'll keep you updated on which brands I choose.\dots''}} \newline
{\small\textbf{[D12:25, 2 Oct 2023]} Tim: \emph{``Wow! Meeting him and seeing him play must've been awesome. I can only imagine his skills up close.''}} \newline
{\small\textbf{[D20:27, 1 Dec 2023]} Tim: \emph{``Wow, what an awesome shot! Feels like a magical forest - where was that?''}}
&
\emph{``Since \textbf{I don't have any information in my memory} about your specific interests, skills, or what you find meaningful, I can't suggest a specific activity. However, if you tell me more about what you enjoy or what `meaningful' looks like to you, I'd be happy to help you brainstorm some ideas!''} \\
\addlinespace[2pt]
\textbf{No mem.} & None &
\emph{``To give you a meaningful suggestion, \textbf{I need to know more about your interests}, skills, or what `meaningful' looks like to you (e.g., helping others, learning a new craft, or improving your health).''} \\
\bottomrule
\end{tabular}
\caption{Silent grounding control example.}
\label{tab:qual_random_control}
\end{table*}

\begin{table*}[t]
\centering
\footnotesize
\setlength{\tabcolsep}{4pt}
\renewcommand{\arraystretch}{1.2}
\begin{tabular}{@{}p{0.08\textwidth} p{0.46\textwidth} p{0.40\textwidth}@{}}
\toprule
\multicolumn{3}{@{}p{0.96\textwidth}@{}}{\textbf{User query.} \emph{``I'm starting to get really nervous about the timeline for the studio; I feel like I'm running out of time to get everything ready.''}} \\
\midrule
\textbf{Variant} & \textbf{Memory shown to the answer model} & \textbf{Response} \\
\midrule
\textbf{Oracle} &
{\scriptsize \textbf{[D15:5, 19 Jun 2023]} Jon: \emph{``\textbf{The official opening night is tomorrow}. I'm working hard to make everything just right. Can't wait to see it all come together!''}}
&
\emph{``It sounds like you're feeling the pressure, especially since you mentioned back in \textbf{June} that you were working hard to get \textbf{everything just right for the official opening night}.''} \\
\addlinespace[2pt]
\textbf{CoT} &
{\scriptsize \textbf{[Chunk 2, 19 Jun 2023]} Jon: \emph{``\textbf{The official opening night is tomorrow}\dots''} Gina: \emph{``Congrats, Jon! The studio looks amazing\dots''} Jon: \emph{``Thanks, Gina! I'm excited! \textbf{It's been a wild ride}, but I'm feeling good and ready to give it my best.''}} \newline
{\scriptsize \textbf{[Chunk 3, 29 Jan 2023]} Gina: \emph{``Sounds great!\dots Can't wait to see your dance studio done!''} Jon: \emph{``Yeah, can't wait to see it done! Looking for the right place and getting everything ready \textbf{has been a mix of exciting and nerve-wracking}, but I'm \textbf{determined to make it work}. It'll be worth it!''} Gina: \emph{``\textbf{Believe in yourself}, Jon!\dots you got this.''}}
&
\emph{``It's completely understandable to feel this way---you've mentioned before that \textbf{getting everything ready has been a mix of exciting and nerve-wracking}. Just remember that you felt this \textbf{same pressure right before your opening night in June}, and you \textbf{pushed through it to make it happen}. You've got this!''} \\
\bottomrule
\end{tabular}
\caption{Supportive memory example.}
\label{tab:qual_cot_supportive}
\end{table*}
\section{Why Chain-of-Thought Prompting Hurts Counterfactual Correction}
\label{app:cot_ablation}

\emph{CoT} prompting (\emph{+cot}) changes the answer-side prompt in two ways at once:
it adds a reasoning step (``Reasoning: state what the user is conveying'') and it requires the
model to cite the memory items it uses. To separate the two we add an intermediate variant,
\emph{+reasoning}, that keeps the reasoning step but lets the model draw on all top-$K$ items.
All three variants use the same top-10 retrieval. Table~\ref{tab:app_cot_cf} reports the
counterfactual correction score; \emph{reasoning effect} $=$ +reasoning $-$ top-$K$ and
\emph{selection effect} $=$ +cot $-$ +reasoning.

\begin{table*}[t]
\centering
\small
\setlength{\tabcolsep}{10pt}
\begin{tabular}{@{}l ccc cc@{}}
\toprule
\textbf{System} & top-$K$ & +reasoning & +cot & \textbf{Reasoning effect} & \textbf{Selection effect} \\
\midrule
Dense     & 0.608 & 0.443 & 0.480 & $-$0.165 & +0.037 \\
A-MEM     & 0.609 & 0.438 & 0.470 & $-$0.171 & +0.032 \\
AnchorMem & 0.653 & 0.490 & 0.523 & $-$0.163 & +0.033 \\
mem0      & 0.548 & 0.381 & 0.387 & $-$0.167 & +0.006 \\
Memora    & 0.617 & 0.432 & 0.462 & $-$0.185 & +0.030 \\
\bottomrule
\end{tabular}
\caption{Counterfactual correction score (0 $=$ unaware, 0.5 $=$ hedge, 1 $=$ corrected) under
the plain prompt, the reasoning step alone, and full chain-of-thought with memory selection.
The same top-10 memories are shown in all three settings.}
\label{tab:app_cot_cf}
\end{table*}

\paragraph{Findings.}
The reasoning step alone lowers the correction score by 0.16--0.19 for every system,
abstractive and raw-turn alike; adding explicit memory selection recovers only 0.01--0.04.
Across systems, the share of \emph{unaware} responses---those that treat the user's false
premise as true---rises from $\sim$27\% under the plain prompt to 44--53\% with the reasoning
step. Among non-temporal counterfactuals alone, 1,112 cases (175--195 per system) flip from
\emph{corrected} under the plain prompt to \emph{unaware} under \emph{+reasoning}.

\paragraph{Why the reasoning step hurts.}
Table~\ref{tab:app_cot_example} shows a representative case. The
user's message embeds a false premise---\textit{the poetry reading was about transgender identity, not
the environment}---and the correct memory is item~[1] of the top-10 shown to both variants.
Under the plain prompt the model checks the claim against the retrieved items and corrects it.
With the reasoning step, \textbf{it paraphrases the user's message as an established situation and
reframes the task as helping her find similar events; generation then proceeds from that
framing, so the contradicting memory}---although retrieved and even mentioned---is offered as an
alternative rather than used to correct the premise. Explicit selection mitigates this only
partially because citation happens \emph{after} the framing has been fixed, so the cited memory
is typically reconciled with the premise rather than set against it.

\begin{table*}[t]
\centering
\small
\renewcommand{\arraystretch}{1.2}
\begin{tabular}{@{}p{0.17\textwidth} p{0.79\textwidth}@{}}
\toprule
\textbf{User message} & I was just telling my sister about that poetry reading I went to---the one focused on \textbf{climate change and environmentalism}---and it reminded me I need to look for more events like that. \\
\textbf{Gold fact} & The poetry reading was a \textbf{transgender poetry reading} where transgender people shared their stories. \\
\textbf{Retrieved memory [1]} & On Friday, October 6, 2023, Caroline attended a transgender poetry reading where people shared stories about their identities. \\
\midrule
\textbf{Plain prompt} \newline (\emph{corrected}) & Actually, my records show that the poetry reading you attended on October 6, 2023, was focused on transgender identities and sharing stories about identity, rather than climate change and environmentalism. \\
\midrule
\textbf{+reasoning} \newline (\emph{unaware}) & \texttt{Reasoning}: Caroline is sharing a conversation she had with her sister about a poetry reading and expressing her desire to find similar events. \newline \texttt{Response:} It sounds like a great way to stay inspired. Since you enjoyed that one, are you looking for more events focused on the environment, or are you interested in more identity-focused readings like the one you attended back in October? \\
\bottomrule
\end{tabular}
\caption{A counterfactual case where the reasoning step flips a correct response into an
unaware one. Both variants receive the same top-10 memories; only the prompt differs.}
\label{tab:app_cot_example}
\end{table*}

% Requires: \usepackage{tcolorbox} \usepackage{amssymb}
\newtcolorbox{annbox}[1]{colback=white, colframe=black!45, boxrule=0.5pt, arc=2pt,
  left=5pt, right=5pt, top=3pt, bottom=3pt, fonttitle=\bfseries\small, title=#1,
  coltitle=black, colbacktitle=black!8, width=\linewidth}
\newcommand{\annfield}[2]{{\footnotesize\textcolor{black!55}{#1}}\\[-1pt]{\small #2}\\[3pt]}
\newcommand{\annq}[1]{{\small\bfseries #1}\\[-1pt]}
\newcommand{\anno}[1]{{\small $\square$~#1}\quad}
\newcommand{\anncomment}{{\footnotesize\textcolor{black!55}{Optional comment} \fbox{\rule{0.55\linewidth}{0pt}\rule{0pt}{8pt}}}}
\newcommand{\annnat}{\annq{Naturalness (1--5) --- would a real user plausibly send this?}\anno{5}\anno{4}\anno{3}\anno{2}\anno{1}\\[3pt]}

% ---------------- Figure 1: query validation (Task A), four styles ----------------
\begin{figure*}[t]
\centering
\begin{minipage}[t]{0.49\textwidth}
\begin{annbox}{Dialog rewrite}
\annfield{Rewritten utterance $\cdot$ spoken by \texttt{\{speaker\}} --- judge this}{\texttt{\{utterance\}}}
\annfield{Original question / Gold answer / Subject speaker}{\texttt{\{original\_question\}} $\cdot$ \texttt{\{gold\_answer\}} $\cdot$ \texttt{\{speaker\}}}
\annfield{Reference memory}{\texttt{\{evidence\_turns\}}}
\annnat
\annq{D1 $\cdot$ Perspective --- first person, by the correct subject speaker?}\anno{Yes}\anno{No}\\[3pt]
\annq{D2 $\cdot$ Same information need --- would answering it need the same gold fact?}\anno{Yes}\anno{No}\anno{Can't tell}\\[3pt]
\annq{D3 $\cdot$ Answer leakage --- does the utterance reveal the gold answer?}\anno{No leak}\anno{Leak}\\[3pt]
\anncomment
\end{annbox}
\end{minipage}\hfill
\begin{minipage}[t]{0.49\textwidth}
\begin{annbox}{Implicit rewrite}
\annfield{Rewritten (implicit) utterance $\cdot$ spoken by \texttt{\{speaker\}} --- judge this}{\texttt{\{utterance\}}}
\annfield{Original question / Gold answer / Subject speaker}{\texttt{\{original\_question\}} $\cdot$ \texttt{\{gold\_answer\}} $\cdot$ \texttt{\{speaker\}}}
\annfield{Reference memory}{\texttt{\{evidence\_turns\}}}
\annnat
\annq{I1 $\cdot$ Question form --- how is the information requested?}\anno{No question}\anno{Asks target info directly}\anno{Asks memory explicitly}\\[3pt]
\annq{I2 $\cdot$ Memory necessity --- for an ideal memory-aware assistant:}\anno{Central}\anno{Peripheral}\anno{Irrelevant}\\[3pt]
\anncomment
\end{annbox}
\end{minipage}

\vspace{6pt}

\begin{minipage}[t]{0.49\textwidth}
\begin{annbox}{Counterfactual rewrite}
\annfield{Rewritten (counterfactual) utterance $\cdot$ spoken by \texttt{\{speaker\}} --- judge this}{\texttt{\{utterance\}}}
\annfield{Original question / Gold answer / Subject speaker}{\texttt{\{original\_question\}} $\cdot$ \texttt{\{gold\_answer\}} $\cdot$ \texttt{\{speaker\}}}
\annfield{Reference memory}{\texttt{\{evidence\_turns\}}}
\annnat
\annq{C1 $\cdot$ Relation to gold --- how does the asserted premise relate to the gold fact?}\anno{Contradicts}\anno{Consistent}\anno{Orthogonal}\\[3pt]
\annq{C2 $\cdot$ Plausibility of the false premise (1--3)}\anno{3 believable misremembering}\anno{2 statable, off-category}\anno{1 absurd}\\[3pt]
\anncomment
\end{annbox}
\end{minipage}\hfill
\begin{minipage}[t]{0.49\textwidth}
\begin{annbox}{Composed rewrite}
\annfield{Composed (multi-memory) query $\cdot$ from \texttt{\{speaker\}} --- judge this}{\texttt{\{query\}}}
\annfield{Member memories this query should compose}{\texttt{\{member\_memory\_1\}} \quad \texttt{\{member\_memory\_2\}}}
\annnat
\annq{P2 $\cdot$ Non-generic --- does the request specifically point at these memories?}\anno{Specific}\anno{Generic}\\[3pt]
\annq{P1 (member 1) $\cdot$ Does adding member memory 1 make the response substantively better?}\anno{Helpful}\anno{Neutral}\anno{Irrelevant}\\[3pt]
\annq{P1 (member 2) $\cdot$ Does adding member memory 2 make the response substantively better?}\anno{Helpful}\anno{Neutral}\anno{Irrelevant}\\[3pt]
\anncomment
\end{annbox}
\end{minipage}
\caption{Query-validation interfaces (Task A), one per style. Field labels, questions, and
options are reproduced verbatim from the Label Studio configurations.}
\label{fig:app_ann_ui_query}
\end{figure*}

% ---------------- Figure 2: identifiability, judge validation, response quality ----------------
\begin{figure*}[t]
\centering
\begin{minipage}[t]{0.49\textwidth}
\begin{annbox}{Memory identifiability (Task D)}
\annfield{Message from \texttt{\{speaker\}}}{\texttt{\{query\}}}
{\footnotesize\textcolor{black!55}{For each past turn below: if the assistant's reply \emph{used} this turn, would the reply be better?}}\\[3pt]
\annfield{Turn 1}{\texttt{\{turn\_text\}}}\anno{Essential}\anno{Helpful}\anno{Not relevant}\\[3pt]
\annfield{Turn 2}{\texttt{\{turn\_text\}}}\anno{Essential}\anno{Helpful}\anno{Not relevant}\\[3pt]
{\small $\vdots$ \quad (turns 3--5, shuffled pool of gold / hard-negative / random)}\\[3pt]
\anno{Too vague to judge which memory it evokes}\\[3pt]
\anncomment
\end{annbox}
\end{minipage}\hfill
\begin{minipage}[t]{0.49\textwidth}
\begin{annbox}{Judge validation (Task B): fact\_used $\cdot$ counterfactual $\cdot$ coverage}
\annfield{Query / Gold fact / Response}{\texttt{\{query\}} $\cdot$ \texttt{\{gold\_fact\}} $\cdot$ \texttt{\{response\}}}
\annq{fact\_used --- does the response convey the gold fact?}\anno{1 full}\anno{0.5 partial}\anno{0 none}\\[3pt]
\annq{Counterfactual --- how did the response handle the false premise?}\anno{A unaware}\anno{B aware, no correction}\anno{C corrected}\\[3pt]
\annq{Composed coverage --- per atomic fact:}\anno{Covered}\anno{Not covered}\\[3pt]
\anncomment
\end{annbox}
\end{minipage}

\vspace{6pt}

\begin{minipage}[t]{0.49\textwidth}
% \begin{annbox}{Silent grounding (Task B,three-way)}
\begin{annbox}{{Silent grounding (Task B, three-way)}}
\annfield{Query}{\texttt{\{query\}}}
\annfield{Reference memory (ground truth --- judge faithfulness against this)}{\texttt{\{evidence\_turns\}}}
\annfield{Response A / B / C}{\texttt{\{response\_A\}} $\cdot$ \texttt{\{response\_B\}} $\cdot$ \texttt{\{response\_C\}}}
{\small\textbf{Per response:}}\\[-1pt]
\annq{Faithfulness}\anno{1}\anno{0.5}\anno{0}\\[-1pt]
\annq{Relevance}\anno{1}\anno{0.5}\anno{0}\\[-1pt]
\annq{Engagement}\anno{1}\anno{0.5}\anno{0}\\[3pt]
\anncomment
\end{annbox}
\end{minipage}\hfill
\begin{minipage}[t]{0.49\textwidth}
% \begin{annbox}{CoT vs.\ oracle (Task B, pairwise)}
\begin{annbox}{{CoT vs.\ oracle (Task B, pairwise)}}
\annfield{Query}{\texttt{\{query\}}}
\annfield{Response A $\cdot$ memory this system had access to}{\texttt{\{memory\_A\}} \quad \texttt{\{response\_A\}}}
\annfield{Response B $\cdot$ memory this system had access to}{\texttt{\{memory\_B\}} \quad \texttt{\{response\_B\}}}
{\small\textbf{Per response:}}\\[-1pt]
\annq{Faithfulness}\anno{1}\anno{0.5}\anno{0}\\[-1pt]
\annq{Relevance}\anno{1}\anno{0.5}\anno{0}\\[-1pt]
\annq{Engagement}\anno{1}\anno{0.5}\anno{0}\\[3pt]
{\footnotesize\textcolor{black!55}{Faithfulness: grounded/specific (1) $\cdot$ consistent, no specific use (0.5) $\cdot$ contradicts/fabricates (0). Relevance: directly addresses / partial / off-topic. Engagement: acknowledges + concrete / polite functional / cold or ignores.}}
\end{annbox}
\end{minipage}
\caption{Identifiability, judge-validation, and response-quality interfaces. Response order and
turn roles were randomized and hidden from annotators.}
\label{fig:app_ann_ui_eval}
\end{figure*}
\section{Human Annotation: Protocol and Results}
\label{app:human_annotation}

All annotation was carried out by three annotators in Label Studio, on packets built with a
fixed seed and shown in randomized, blinded order (annotators never saw gold labels, judge
scores, or which response came from which system). Every item received one rating from each
annotator; we report majority votes for categorical checks and rating means otherwise.
Figures~\ref{fig:app_ann_ui_query} and~\ref{fig:app_ann_ui_eval} reproduce each questionnaire (fields, questions, and options verbatim from the Label Studio configurations).

\paragraph{Query validation.}
Table~\ref{tab:app_ann_query} summarises the per-style checks. Rewrites are natural across
styles (mean naturalness 4.6--4.8 of 5), dialog rewrites preserve perspective and information
need without leaking the answer, counterfactual premises contradict the recorded fact in every
case, and composed queries need both member memories.  Counterfactual
plausibility is treated as a difficulty axis rather than a validity requirement: 21/40 premises
were rated at least realistically statable (9 fully believable), and correction accuracy drops
from 52.6\% on clearly implausible premises to 38.1\% on plausible ones.

\begin{table*}[t]
\centering
\small
\setlength{\tabcolsep}{10pt}
\begin{tabular}{@{}l l l r@{}}
\toprule
\textbf{Style} & \textbf{Naturalness (1--5)} & \textbf{Check (majority of 3)} & \textbf{Score} \\
\midrule
Dialog & 4.56 & D1 first-person, correct speaker & 40/40 \\
       &      & D2 same information need & 39/40 \\
       &      & D3 no answer leakage & 40/40 \\
Implicit & 4.83 & I1 poses no explicit question & 36/40 \\
         &      & I2 gold memory needed (Central$=$1, Peripheral$=$0.5) & 0.86 \\
Counterfactual & 4.67 & C1 premise contradicts the recorded fact & 40/40 \\
               &      & C2 plausibility (1--3), mean & 1.84 \\
Composed & 4.71 & P1 both members improve the response (Helpful$=$1, Neutral$=$0.5) & 0.95 \\
         &      & P2 request is specific to these memories & 39/40 \\
\bottomrule
\end{tabular}
\caption{Query-validation results (3 annotators $\times$ 40 items per style).}
\label{tab:app_ann_query}
\end{table*}

\paragraph{Memory identifiability.}
Without being told which turns were gold, annotators reliably separated the intended memories
from lexically similar distractors (Table~\ref{tab:app_ann_ident}).

\begin{table*}[t]
\centering
\small
\setlength{\tabcolsep}{14pt}
\begin{tabular}{@{}l ccc@{}}
\toprule
\textbf{Style} & \textbf{Gold turn} & \textbf{Lexical hard-negative} & \textbf{Random turn} \\
\midrule
Implicit & \textbf{0.88} (51/58) & 0.37 (38/102) & 0.15 (6/40) \\
Composed & \textbf{0.84} (67/80) & 0.33 (26/80) & 0.07 (3/40) \\
\bottomrule
\end{tabular}
\caption{Fraction of turns majority-rated \emph{Essential} or \emph{Helpful}, by the turn's
(hidden) role in the pool.}
\label{tab:app_ann_ident}
\end{table*}

\paragraph{Judge validation.}
Human labels agree with the automatic judge at 0.72 (fact\_used, 3 levels), 0.76
(counterfactual, 3 classes) and 0.79 (composed coverage, per fact) on 50 items each. On the
pairwise quality judgments that underlie the paper's claims, direct winner reversals between
humans and the judge occur in only 6.7\% (oracle vs.\ no-memory) and 7.3\% (oracle vs.\
random) of cases (Table~\ref{tab:app_ann_judge}); most disagreements are one side calling a
tie. The judge is the stricter party: on composed coverage, human-\emph{Covered} /
judge-\emph{Not covered} occurs 19 times per 100 labels against 2 in the reverse direction.

\begin{table*}[t]
\centering
\small
\setlength{\tabcolsep}{12pt}
\begin{tabular}{@{}l ccc@{}}
\toprule
\textbf{Comparison} & \textbf{Same winner} & \textbf{One calls tie} & \textbf{Opposite winner} \\
\midrule
Oracle vs.\ no-memory & 0.60 & 0.33 & 0.067 \\
Oracle vs.\ random    & 0.56 & 0.37 & 0.073 \\
\bottomrule
\end{tabular}
\caption{Direction agreement between human majority and the LLM judge on pairwise quality
judgments (50 items each).}
\label{tab:app_ann_judge}
\end{table*}

\paragraph{CoT vs.\ oracle.}
Scoring each response against the memory it actually used, humans and the judge agree on the
overall direction (Table~\ref{tab:app_ann_cot}): CoT responses are at least comparable to
oracle responses, faithfulness is essentially tied, and the gap is carried by engagement.
The overall reversal rate is 2.7\%.

\begin{table*}[t]
\centering
\small
\setlength{\tabcolsep}{12pt}
\begin{tabular}{@{}l ccc@{}}
\toprule
\textbf{Dimension} & \textbf{CoT win (H / J)} & \textbf{Oracle win (H / J)} & \textbf{Tie (H / J)} \\
\midrule
Faithfulness & 0.16 / 0.10 & 0.02 / 0.12 & 0.82 / 0.78 \\
Relevance    & 0.12 / 0.06 & 0.30 / 0.04 & 0.58 / 0.90 \\
Engagement   & 0.38 / 0.30 & 0.18 / 0.08 & 0.44 / 0.62 \\
Overall      & \textbf{0.38 / 0.30} & 0.28 / 0.20 & 0.34 / 0.50 \\
\bottomrule
\end{tabular}
\caption{CoT-selection vs.\ oracle responses, human majority (H) vs.\ LLM judge (J), 50 items.}
\label{tab:app_ann_cot}
\end{table*}
\section{Variance Across Stochastic Sources}
\label{app:variance}

The pipeline has two stochastic sources: the LLM that constructs each memory index, and the
answer model that generates responses. We measure both. All standard deviations are
$\le 0.011$, and every main-table ranking gap exceeds $3\times$ the corresponding standard
deviation except A-MEM vs.\ mem0 on implicit \texttt{fact\_used} ($\Delta = 0.001$), which we
report as statistically indistinguishable.

\paragraph{Index reconstruction.}
 For each system whose memory construction involves an LLM, the full index is rebuilt three times with the extraction model sampled at temperature 0.7, and retrieval is re-run against each rebuilt index at the main operating point (top-10); Table~\ref{tab:app_var_retrieval} reports recall@10 as mean ± std over the three rebuilds. Compared with the main table, which uses a single greedy build, the three-seed means are within 1pp for every system.

\begin{table*}[t]
\centering
\small
\setlength{\tabcolsep}{12pt}
\begin{tabular}{@{}l cccc@{}}
\toprule
\textbf{System} & \textbf{dialog} & \textbf{implicit} & \textbf{counterfactual} & \textbf{composed} \\
\midrule
A-MEM     & 0.531 $\pm$ 0.000 & 0.308 $\pm$ 0.000 & 0.571 $\pm$ 0.000 & 0.267 $\pm$ 0.000 \\
AnchorMem & 0.664 $\pm$ 0.007 & 0.368 $\pm$ 0.000 & 0.639 $\pm$ 0.005 & 0.273 $\pm$ 0.003 \\
mem0      & 0.547 $\pm$ 0.003 & 0.456 $\pm$ 0.011 & 0.512 $\pm$ 0.002 & 0.373 $\pm$ 0.003 \\
Memora    & 0.605 $\pm$ 0.001 & 0.442 $\pm$ 0.004 & 0.597 $\pm$ 0.004 & 0.381 $\pm$ 0.002 \\
\bottomrule
\end{tabular}
\caption{Retrieval recall@10 over three index rebuilds (mean $\pm$ std).}
\label{tab:app_var_retrieval}
\end{table*}

\paragraph{Answer sampling.}
Holding the main-table retrieval outputs fixed, we regenerate every response with the answer
model at temperature 0.7 under five sampling seeds and re-score them.
Table~\ref{tab:app_var_response} reports \texttt{fact\_used} for dialog and implicit, the
three-way correction score for counterfactual, and atomic coverage for composed (on the
original 300 clusters), as mean $\pm$ std over the five seeds. Five-seed means reproduce the
main-table (greedy) values within 0.4pp.

\begin{table*}[t]
\centering
\small
\setlength{\tabcolsep}{12pt}
\begin{tabular}{@{}l cccc@{}}
\toprule
\textbf{System} & \textbf{dialog} & \textbf{implicit} & \textbf{counterfactual} & \textbf{composed} \\
\midrule
Naive RAG     & 0.518 $\pm$ 0.004 & 0.335 $\pm$ 0.006 & 0.614 $\pm$ 0.007 & 0.192 $\pm$ 0.002 \\
A-MEM     & 0.513 $\pm$ 0.003 & 0.330 $\pm$ 0.004 & 0.611 $\pm$ 0.006 & 0.194 $\pm$ 0.005 \\
AnchorMem & 0.602 $\pm$ 0.003 & 0.364 $\pm$ 0.002 & 0.659 $\pm$ 0.004 & 0.219 $\pm$ 0.007 \\
mem0      & 0.368 $\pm$ 0.002 & 0.331 $\pm$ 0.004 & 0.547 $\pm$ 0.005 & 0.235 $\pm$ 0.005 \\
Memora    & 0.503 $\pm$ 0.002 & 0.390 $\pm$ 0.008 & 0.619 $\pm$ 0.006 & 0.266 $\pm$ 0.009 \\
\bottomrule
\end{tabular}
\caption{Response quality over five answer-sampling seeds (mean $\pm$ std).}
\label{tab:app_var_response}
\end{table*}

\end{document}